\documentclass[a4paper]{styles/svproc}
\usepackage[english]{babel}
\usepackage[utf8]{inputenc}

\usepackage{color,xcolor,ucs}

\usepackage{subcaption}
\usepackage{floatrow}
\usepackage{tabularx}
\usepackage{float}
\usepackage{amsfonts}
\usepackage{helvet}         
\usepackage{courier}        
\usepackage{type1cm}        
\usepackage{amsmath}
\usepackage{amssymb}
\usepackage{makeidx}         
\usepackage{comment}         
\usepackage{graphicx}        
\usepackage{multicol}        
\usepackage[bottom]{footmisc}
\usepackage{bm}
\usepackage{bbm}

\usepackage{cite}
\usepackage{url}

\usepackage[unicode=true, bookmarks=true,colorlinks=true]{hyperref}
\usepackage{xr-hyper}

\usepackage{algpseudocode,algorithm,algorithmicx}

\usepackage{xspace}
\usepackage{rotating}

\usepackage{tikz}
\usepackage{tkz-graph}
\usetikzlibrary{arrows,shapes,shadows,positioning,calc}

\usepackage{graphicx}
\usepackage{threeparttable}
\usepackage{multirow}
\usepackage[font=scriptsize,labelfont=bf]{caption}

\usepackage{cancel}

\usepackage{url}
\usepackage{wrapfig}
\usepackage{booktabs}
\usepackage{multirow}

\usepackage{xr-hyper}
\usepackage{hyperref}

\DeclareMathOperator*{\argmax}{arg\,max}

\newcommand{\prob}[1]{\ensuremath{\mathbb{P}({#1})}}

\algrenewcommand\algorithmicrequire{\textbf{Input:}}
\algrenewcommand\algorithmicensure{\textbf{Input:}}
\algnewcommand{\LineComment}[1]{\State \(\triangleright\) #1}

\def\CX{\mathcal{X}} 
\def\CA{\mathcal{A}}
\def\aSpace{\mathcal{A}}
\def\CZ{\mathcal{Z}}
\def\Zspace{\mathcal{Z}}

\def\Ospace{\mathcal{O}}
\def\gobs{\Bar{z}}
\def\gobSpace{\Bar{\Zspace}}

\def\simpfunc{\beta}
\def\tb{{b}}
\def\shis{h}

\newcommand{\exptsimp}[2]{\ensuremath{\underset{{#1}}{\mathbb E^{#2}}}}
\newcommand{\emathbb}[1]{\ensuremath{\underset{{#1}}{\mathbb E}}}
\def\topo{\tau}
\def\indicator{\mathbbm{1}}
\def\zchild{\gamma^{\topo}}

\def\action{a} 

\def\policy{\pi}
\def\bpolicy{\bar{\policy}} 
\def\bpolicystar{\bar{\policy}^{+}} 

\def\policyc{\policy^{\topo_C}}

\def\BelTree{\mathbb{T}}

\newcommand{\probz}[1]{\ensuremath{\mathbb{P}_{Z}({#1})}}
\newcommand{\probo}[1]{\ensuremath{\mathbb{P}_{O}({#1})}}
\newcommand{\probg}[1]{\ensuremath{\mathbb{P}_{\bar{Z}}({#1})}}

\begin{document}
\mainmatter              
\title{
Simplified POMDP Planning with an Alternative Observation Space and Formal Performance Guarantees}

\titlerunning{Simplified POMDP Planning with an Alternative Observation Space}  
%
\author{Da Kong\inst{1} \and Vadim Indelman\inst{2}}
%

%
\institute{Technion Autonomous Systems Program \\
	\and
	Department of Aerospace Engineering \\
	Technion - Israel Institute of Technology, Haifa 32000, Israel \\
	\email{da-kong@campus.technion.ac.il}, \email{vadim.indelman@technion.ac.il}
}
\maketitle              

\begin{abstract}
\vspace{-10pt}
Online planning under uncertainty in partially observable domains is an essential capability in robotics and AI. The partially observable Markov decision process (POMDP)  is a mathematically principled framework for addressing decision-making problems in this challenging setting. However, finding an optimal solution for POMDPs is computationally expensive and is feasible only for small problems. In this work, we contribute a novel method to simplify POMDPs by switching to an alternative, more compact, observation space and simplified model to speedup planning with formal performance guarantees. We introduce the notion of belief tree topology, which encodes the levels and branches in the tree that use the original and alternative observation space and models. Each belief tree topology comes with its own policy space and planning performance. Our key contribution is to derive bounds between the optimal Q-function of the original POMDP and the simplified tree defined by a given topology with a corresponding simplified policy space. These bounds are then used as an adaptation mechanism between different tree topologies until the optimal action of the original POMDP can be determined. Further, we consider a specific instantiation of our framework, where the alternative observation space and model correspond to a setting where the state is fully observable.   
We evaluate our approach in simulation, considering exact and approximate POMDP solvers and demonstrating a significant speedup while preserving solution quality. We believe this work opens new exciting avenues for online POMDP planning with formal performance guarantees.

\end{abstract}

\section{Introduction}
Decision-making under uncertainty in partially observable domains is a fundamental problem in robotics and AI. 	A key required capability is to operate autonomously online in a partially observable setting, where the	agent maintains a probability distribution (belief) over the state.  	The partially observable Markov decision process (POMDP) \cite{Kaelbling98ai} is a mathematically principled framework for addressing decision-making problems in these challenging settings.
By considering all kinds of uncertainty and planning in the belief space, POMDP has shown proven advantages in general decision-making problems under uncertainty and many robotics tasks.

\begin{wrapfigure}{t}{0.45\textwidth}
	\vspace{-15pt}
	\centering
	\includegraphics[width=\textwidth]{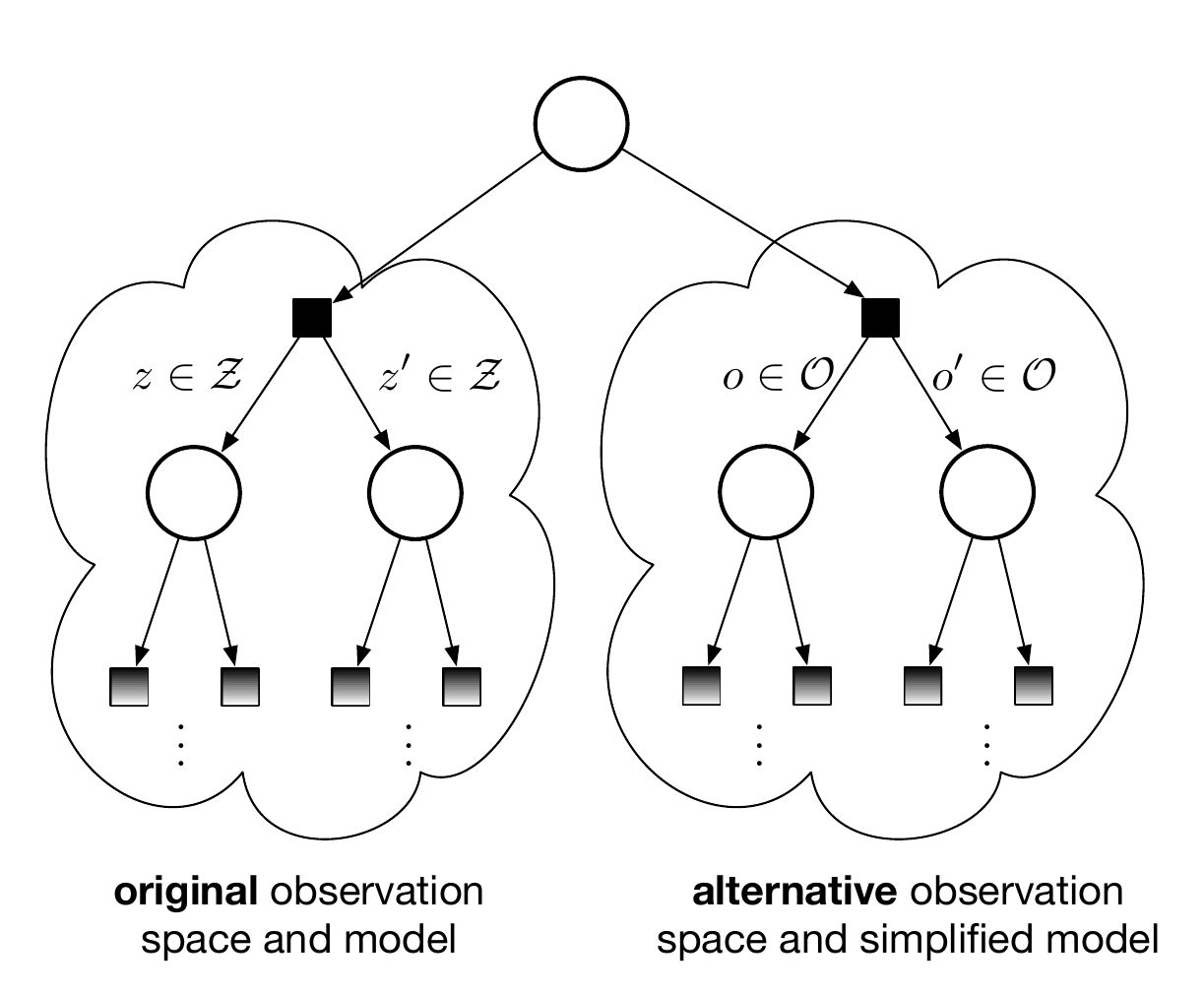}
	\caption{The idea of using alternative observation space and model to simplify POMDP.}
	\label{fig:idea}
	\vspace{-5pt}
\end{wrapfigure}

However, deriving the optimal solution for a general POMDP is computationally infeasible due to its inherent complexity, attributed to the \textit{curse of dimensionality} and the \textit{curse of history}. 
As a result, practical online methods often resort to approximating the full POMDP using various techniques \cite{Roy05jair, Silver10nips, Sunberg18icaps}.
Notable approximate solvers include POMCP \cite{Silver10nips}, employing Monte Carlo rollouts, and DESPOT \cite{Ye17jair, Somani13nips}, which leverages branch-and-bound and dynamic programming techniques.
Recent developments have also introduced methods for approximating the information state \cite{Subramanian22jmlr}, as well as using a finite memory window \cite{Kara22jmlr}.

In addition to approximation, recent research has focused on simplifying POMDPs while providing formal performance guarantees. 
For example, these prior studies encompass simplifying the observation model \cite{LevYehudi24aaai}, reducing the state and observation space \cite{Barenboim23nips}, sparsifying beliefs \cite{Elimelech22ijrr}, and employing multi-level simplification strategies \cite{Zhitnikov24ijrr, Hoerger23ijrr}. Kara et al. ~\cite{Kara22jmlr} achieved POMDP simplification by simplifying historical memory feedback and demonstrated near optimality. Additionally, Flaspohler et al. \cite{Flaspohler20nips} proposed an online method for generating macro actions to support open-loop planning across multiple steps with performance guarantees, albeit necessitating the expensive calculation of the Value of Information for observations.

From the practical side, simplifying the observation space and model is crucial in numerous visual tasks, especially in the context of active visual SLAM (see e.g.~\cite{Giuliari21iros}). Moreover, for safety-critical robotics and AI tasks, it is essential to provide a rigorous theoretical analysis that the simplified models can effectively represent the original POMDP and establish a performance guarantee. This involves ensuring that the simplified objective or value function has a bounded error compared to the original. Some simplification methods can offer theoretical performance guarantees, demonstrating a bounded value function error when comparing the original POMDP and the simplified model \cite{Barenboim23nips, LevYehudi24aaai}, as well as between the theoretical and estimated models \cite{Lim23jair}. Given such performance guarantees, the simplified POMDP solver can adapt the simplification level to ensure the same optimal policy is calculated as the original one. However,  current methods do not consider adapting the simplified observation space and model simultaneously. Moreover, Lev-Yeudi et al.~\cite{LevYehudi24aaai} focus solely on simplifying the observation model within the original complex space.

In this paper, we introduce a new methodology to speedup POMDP planning with formal performance guarantees by switching to an alternative observation space and simplified model. The alternative observation space may be entirely different than the original observation space. For instance, it could correspond to the space of images with lower resolution, or to a learned latent representation space. The simplified observation model could correspond, e.g.~to a smaller deep neural network. We introduce a novel structure of a simplified belief tree where different levels and branches may use either the original or the alternative observation space and model (see  Fig.~\ref{fig:idea}). We refer to a particular such choice as a belief tree topology.

Each belief tree topology comes with its own policy space and planning performance, which comes with a reduced computational cost, compared to the original POMDP, because of the selective switch to the alternative observation space and simplified model. Further, we consider a specific instantiation of our framework, where the alternative observation space and model correspond to a setting where the state is fully observable. We derive novel bounds for this setting  
between the optimal Q-function of the original POMDP and the simplified POMDP for a given topology, considering the corresponding simplified policy space. These bounds are then used as an adaptation mechanism between different tree topologies until the optimal policy of the original POMDP is determined. 
Finally, we introduce a practical sparse sampling estimator to the proposed simplification and demonstrate that effective estimation can be used to significantly accelerate planning while preserving the solution quality.

To summarize, in this paper we make the following main contributions: (a) We propose a novel adaptive simplified belief tree to switch to alternative observation space and model simultaneously at selected nodes  in the tree. To our knowledge, this work is the first to address simplification of POMDP by an adaptive switching to an alternative observation space. We show that our method also simplifies the policy space, which is of independent interest. (b) We develop a specific instance of an alternative observation space and model that corresponds to full observability, and derive novel bounds that  serve as formal performance guarantees and for adaptation between different topologies. (c) We introduce a practical sparse sampling based estimator of our method. (d) We evaluate our approach in simulation and show it leads to a substatinoal speedup without sacrificing planning performance. This paper is accompanied by supplementary material \cite{Kong24isrr_supplementary} that provides proofs and further details.

\section{Preliminaries and Notations}

The basic model of POMDP is defined as a tuple 
$\langle \CX, \CA, \Zspace, \mathbb{P}_{T}, \mathbb{P}_{Z}, b_k, r \rangle$, where $\CX$ is the state space, $\CA$ is the action space, $\Zspace$ is the observation space. 
The transition model (or motion model) is defined as $\mathbb{P}_{T}(x_{k+1}|x_k, a_k)$, which describes the probabilistic transition of the state from $x_k\in\CX$ to $x_{k+1}\in\CX$ under a certain action $a_k \in \CA$. The observation model is defined as $\probz{z_k|x_k}$, which describes the probability of observation $z_k\in\Zspace$ given a certain state $x_k\in\CX$.
The reward function is considered to be state dependent, $r: \mathcal{X},\mathcal{A} \mapsto \mathbb{R}$, and holds the bounded reward assumption: $r \in [-R_{\max}, R_{\max}]$. 

Given that the true state is uncertain, a belief is maintained to represent the distribution of the current state with regard to history.
The belief at time instant $k$ is defined as $b_k \triangleq \prob{x_{k}| h_{k}} $, where $h_{k} \triangleq \{z_{1:k},a_{0:k-1}\}$ is the history until that time. A propagated history without the latest observation is defined as $h_k^{-} \triangleq \{z_{1:k-1},a_{0:k-1}\}$, and the corresponding propagated belief is $b^-_k \triangleq \prob{x_k|h^-_k}$. 


A policy function is defined as $\pi:\mathcal{H}\to\CA$, which decides actions based on the history. The value function for a certain policy $\pi$ over the planning horizon $L$ is defined as the summation of all expected rewards,  
$V^\pi(b_k)= r[b_k, \pi_k(b_k)] + \sum_{i=k+1}^{   L}\emathbb{z_{k+1:i}}r[b_i,\pi_i(b_i)]$, where for simplicity, in this work we use history and the corresponding belief interchangeably (i.e.~$\pi_i(h_i)\equiv \pi_i(b_i)$, where $b_i=\prob{x_i\mid h_i}$).


The goal of a POMDP is to find the optimal policy $\pi^*$ that maximizes the value function. The optimal value function can be calculated recursively by the Bellman optimality as: $
	V^{\pi*}(b_k) = \max_{a_k} \Big[ r(b_k, a_k) + \emathbb{z_{k+1}} V^{\pi*}(b_{k+1})\Big].
$

\section{Adaptive Observation Belief Tree}

We shall consider an alternative observation space $\mathcal{O}$  instead of the original observation space $\mathcal{Z}$. For instance, this could correspond to the space of images with  a smaller resolution  or to a learned latent vector space. Further, we shall consider a corresponding  simplified observation model $\probo{o \mid x}$, which can be computationally easier to query for likelihood evaluation and to sample than the original observation model $\probz{z \mid x}$. For instance, the original and simplified models could be represented by neural networks (e.g.~\cite{Jonschkowski18corl}) that differ in their architecture, e.g.~large network versus shallower network. We note the simplified observation model can be also defined over the original observation space, as in \cite{LevYehudi24aaai}.

While constructing a belief tree, we may choose to switch to an alternative space and simplified model only at certain levels and branches of the tree. Each such belief tree corresponds to its own planning performance and computational complexity. \emph{How can we decide online where these simplified representations should be used while constructing the belief tree online while providing formal performance guarantees? How can we adaptively transition between the different possibilities?} To address these questions, we first introduce the notion of Adaptive Observation Topology Belief Trees.

\subsection{Definition of Adaptive Observation Topology Belief Tree}
\label{sec:AdatpiveObservationTree}

We use the same definition of posterior belief $b$ and propagated belief $b^-$ as the commom POMDP.
Considering a given belief tree $\BelTree^{\topo}$ where some belief nodes have the alternative observation model and space, 
we define the corresponding topology $\topo$ as follows. The topology $\topo$ is defined in terms of binary variables $\simpfunc^{\topo}(h^{\topo-}_t)\in \{0,1\}$ that indicate for each belief node $b^{\topo-}_t$  with the corresponding propagated history $h^{\tau-}_t$ in the belief tree $\BelTree^{\tau}$ whether we consider as its children an alternative observation space $\mathcal{O}$ and model $\probo{o\mid x}$, or the original observation space $\mathcal{Z}$  and model $\probz{z\mid x}$.
Specifically, if $\simpfunc^{\topo}(h^{\topo-}_t)=1$, then the node $b^{\topo-}_t$ in $\BelTree^{\topo}$ has the original full observation space.
Thus, each belief node $b^{\tau}_t=\prob{x_t| h^{\topo}_t}$ in $\BelTree^{\topo}$ is conditioned on a mix of original and simplified observations, that are part of the corresponding history $h^{\topo}_t$.  See a conceptual illustration in Fig.\ref{fig:idea}.

Further, 
we now define  Bayesian belief update operators that correspond to the original and alternative observation space and model,
	\begin{align*}
		\psi_z(b^{\topo}_{t-1},a_{t-1},z_t) \!\! \triangleq  \! \eta^{-1}_z  b^{\topo-}_{t}(x_{t}) \probz{z_t | x_t}, \  \!
		\psi_o(b^{\topo}_{t-1},a_{t-1},o_t) \!\! \triangleq  \! \eta^{-1}_o  b^{\topo-}_{t}(x_{t}) \probo{o_t | x_t}, 
	\end{align*}
where $b^{\topo-}_{t}(x_{t})  \triangleq\int_{x_{t-1}} b^{\topo}_{t-1}(x_{t-1})\mathbb{P}_{T}(x_{t}|x_{t-1}, a_{t-1}) dx_{t-1}$, and 
\begin{align}
	\eta_z  \triangleq \int_{x_t} \probz{z_t | x_t}  b^{\topo-}_{t}(x_{t}) dx_t, \  \text{and} \ \eta_o  \triangleq \int_{x_t} \probo{o_t | x_t}  b^{\topo-}_{t}(x_{t}) dx_t,
\end{align}
are the normalization constants.
Finally, we define an augmented belief update operator,
\begin{align}
	\psi_{\bar{z}_{t+1}}(b^{\topo}_{t}, a_t, \bar{z}_{t+1}, h^{\topo-}_t) \! \triangleq \!  \simpfunc^{\topo}(h^{\topo-}_{t})\psi_z(b^{\topo}_{t},a_{t},z_{t+1}) + \big(1-\simpfunc^{\topo}(h^{\topo-}_t)\big) \psi_o(b^{\topo}_{t},a_{t},o_{t+1}).
	\label{eq:AugBelUpd}
\end{align}
%
%
%
%
%
%
%
We now define a child node propagation process from $h_t^{\topo-}$ to the set of possible posterior histories at time instant $t\geq1$.
\begin{equation}
	\zchild(h^{\tau-}_t)=\left \{ 
	\begin{array}{ll}
		\{h^{\topo}_t:  h^{\tau}_t = (h^{\topo-}_t, z_t) \ \ \ \forall z_t \in \mathcal{Z} \}, & \text{if }  \beta^{\topo}(h^{\topo-}_t)=1, \\
		\{h^{\topo}_t:  h^{\tau}_t = (h^{\topo-}_t,o_t) \ \ \ \forall o_t \in \mathcal{O} \}, & \text{if } \simpfunc^{\topo}(h^{\topo-}_t)=0. 
	\end{array}
	\right .
\end{equation}
We can now construct the sets $\mathcal{H}^{\topo}_t$ and $\mathcal{H}^{\topo-}_t$, that represent all the possible posterior and propagated  histories, respectively,  from $t=1$ to the end of the planning horizon $t=L$ for a given topology $\topo$,
\begin{align}
	\mathcal{H}^{\topo}_{t} &= \Big\{  h^{\topo}_t: h^{\topo}_t \in \zchild(h^{\tau-}_t), \forall h^{\tau-}_t\in \mathcal{H}^{\tau-}_{t}\Big\},
	\label{def:h_t_set}
	\\
	\mathcal{H}^{\tau-}_{t} &= \Big\{  h^{\tau-}_t: h^{\tau-}_t=(h^{\tau-}_{t-1}, a_{t-1}), \forall a_{t-1}\in \mathcal{A},  \forall h^{\tau}_{t-1} \in \mathcal{H}^{\tau}_{t-1}\Big\},	\label{def:h_t_minus_set}
\end{align}
and $\mathcal{H}^{\tau}_0=\{b_0\}$. These history sets represent all the belief nodes inside the belief tree $\BelTree^{\topo}$. We shall use the set $\mathcal{H}^{\topo-}_{0:L-1}$ to represent all the propagated belief nodes in the belief tree $\BelTree^{\topo}$ from time instant $0$ to $L-1$.

The original full POMDP belief tree can be viewed as a particular case where no belief node has a simplified observation model and space. The original topology is denoted as $\topo_Z$, and the belief tree is denoted as $\BelTree^{\topo_Z}$.
On the other extreme, if all the belief nodes have a simplified observation model and space, we denote the topology as $\topo_O$ and its belief tree as $\BelTree^{\topo_O}$.

Further, let us define an augmented  observation space $\gobSpace$ for a certain topology $\topo$ as a function of a propagated history $h^{\topo-}_t$ and the corresponding binary variable $\simpfunc^{\topo}(h^{\topo-}_t)$ as
 \begin{align}
	 \gobSpace_t(h^{\topo-}_t, \topo) &\triangleq \left \{ 
	\begin{array}{lr}
		\Ospace_t, & \text{if } \simpfunc^{\topo}(h^{\topo-}_t)=0, \\
		\Zspace_t, & \text{if } \simpfunc^{\topo}(h^{\topo-}_t)=1. 
	\end{array}
	\right .
	\label{equ:adaptive-obs-space}
	\end{align}
Based on this 	augmented observation space, we define a corresponding augmented observation model for any $\gobs_t\in\gobSpace_t$,
\begin{align}
	\probg{\gobs_t|x_t, h^{\topo-}_t, \topo} &\triangleq \simpfunc^{\topo}(h^{\topo-}_t)\probz{\gobs_t|x_t} + \big(1-\simpfunc^{\topo}(h^{\topo-}_t)\big) \probo{\gobs_t|x_t}.
	\label{equ:adaptive-obs-model}
 \end{align}
%
%
%
%
We now introduce the notion of a  \textit{topology-dependent policy space}. 
The action $\action_t$ is decided by a policy $\policy^{\topo}_t$ at each node: $\action_t = \policy_t^{\topo}(h^{\topo}_t)$. 
It depends on the history within the history space $\mathcal{H}^{\topo}_t$, which is determined by the given topology $\topo$. Different topologies can lead to different history space and thus to a different policy space. 
For a specific topology $\topo$, the topology-dependent policy space $\Pi^{\topo}$ is the set of all the possible policies that may be adopted,
\begin{equation}
	\Pi^{\topo} \triangleq\{\pi_t^{\topo}: \mathcal{H}^{\topo}_t\mapsto\CA  , 0 \leq t \leq L \}.
	\label{eq:PolicySpace}
\end{equation}
%
The optimal value function for a given topology $\topo$ can be calculated recursively by the Bellman's principle of optimality, i.e.~for any belief $b^{\topo}_{t}$:
	\begin{equation}
		V^{\topo*}_{}(b^{\topo}_t) = \max_{\action_t} \Big[ r(b^{\topo}_t, \action_t) + \exptsimp{\gobs_{t+1}| b^{\topo}_t,a_t}{\topo} V_{}^{\topo*}(\psi_{\bar{z}_{t+1}}(b^{\topo}_{t}, a_t, \bar{z}_{t+1}, h^{\topo-}_{t+1}))\Big ],
	\end{equation}			
%
where we define,  $\exptsimp{\gobs_{t+1}| b^{\topo}_t,a_t}{\topo}\equiv \exptsimp{\gobs_{t+1}| h^{\topo-}_{t+1}}{\topo}=\emathbb{x_t|h^{\topo-}_t}\exptsimp{\gobs_t|x_t, h^{\topo-}_t}{\topo}$, where $\exptsimp{\gobs_t|x_t, h^{\topo-}_t}{\topo}$ is an expectation over $\gobs_t$ with respect to the augmented observation model \eqref{equ:adaptive-obs-model}. Specifically, recalling the augmented belief update operator \eqref{eq:AugBelUpd}, for any function $f(.)$, 
	 \begin{small}	 
			\begin{align*}
			\exptsimp{\gobs_{t}|h^{\topo-}_{t}}{\topo} f(\psi_{\bar{z}}(b^{\topo}_{t-1}, a_{t-1}, \bar{z}_{t}, h^{\topo-}_{t}))&= 
			 \!\!\! \emathbb{x_t|h^{\topo-}_t}\Big[\simpfunc^{\topo}(h^{\topo-}_t)\int_{z_t \in \Zspace_t}\!\!\!\!\!\!\! \probz{z_t|x_t}f(\psi_{z}(b^{\topo}_{t-1}, a_{t-1}, z_{t}))dz_t +\\
			& \big(1-\simpfunc^{\topo}(h^{\topo-}_t)\big)\int_{o_t \in \Ospace_t} \!\!\!\!\!\!\! \probo{o_t|x_t} f(\psi_{o}(b^{\topo}_{t-1}, a_{t-1}, o_{t}))do_t \Big]. \nonumber
		\end{align*}
\end{small}	
%
%
%
%
In this work, we switch the observation space adaptively at some of the belief nodes. This process corresponds to different topologies, each with its own policy space \eqref{eq:PolicySpace}.  We will explore different topologies $\topo_{1},\ldots, \topo_n $, which can be seen as different levels of simplification for POMDP, to speedup planning while providing formal performance guarantees.


\subsection{Performance Guarantees}
\label{subsec:performance-guarantee}
Generally, each topology $\topo$ corresponds to its own planning performance. In this section, we revisit general bounds between the optimal Q-function of the original POMDP, and the simplified POMDP considering some given topology $\topo$, with the corresponding theoretical belief trees  $\BelTree^{\topo_Z}$ and $\BelTree^{\topo}$. These lightweight bounds can then be utilized for planning and for the adaptation between different topologies, as described next.

Specifically, we would like to bound
\begin{equation}
	\big|Q^{\policy^\topo}_{\topo}(b_k,\action_k)-Q^{\policy^{\topo_Z*}}_{\topo_Z}(b_k,\action_k)\big| \leq B(\topo, \policy^\topo, b_k, a_k),
	\label{eq:Boundgeneral}
\end{equation}
where $\policy^{\topo_Z*}$ is the optimal policy of the original POMDP, and $\policy^\topo \in \Pi^{\topo}$ is some policy of the simplified POMDP considering the topology $\tau$.

We can therefore bound the optimal Q-function of the original POMDP as
$
		lb(\topo, \policy^\topo, b_k, a_k) \leq Q^{\policy^{\topo_Z*}}_{\topo_Z}(b_k,\action_k)  \leq  ub(\topo, \policy^\topo, b_k, a_k),
$
where $lb(\topo, \policy^\topo, b_k, a_k) \triangleq Q^{\policy^\topo}_{\topo}(b_k,\action_k) - B(\topo, \policy^\topo, b_k, a_k)$ and $ub(\topo, \policy^\topo, b_k, a_k) \triangleq Q^{\policy^\topo}_{\topo}(b_k,\action_k) +  B(\topo, \policy^\topo, b_k, a_k)$. 

\begin{figure}[h!] 
	\centering 
	\begin{subfigure}[b]{0.45\textwidth} 
		\includegraphics[width=\linewidth]{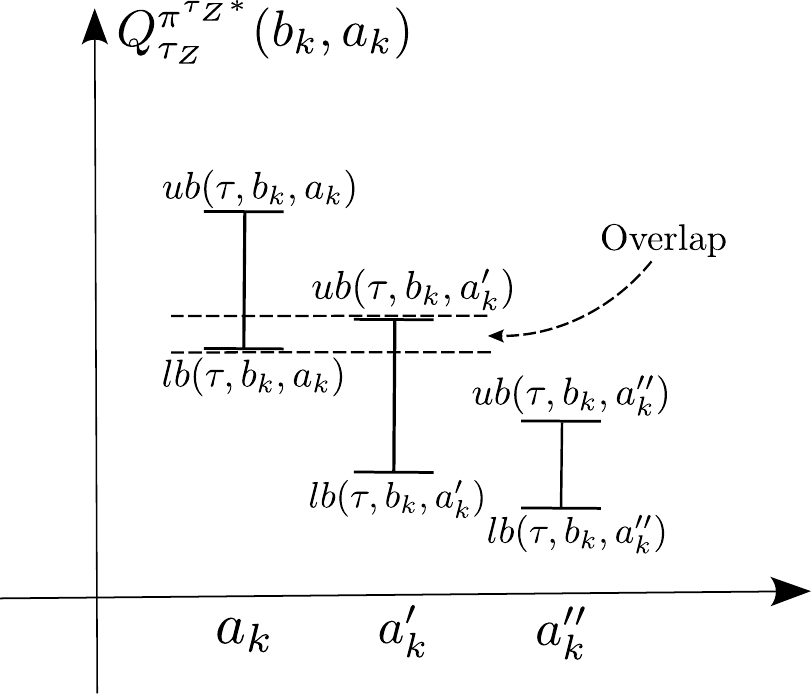}
		\caption{Overlap appears for topology $\topo$}
		\label{fig:guarantee_a}
	\end{subfigure}
	\hfill 
	\begin{subfigure}[b]{0.45\textwidth} 
		\includegraphics[width=\linewidth]{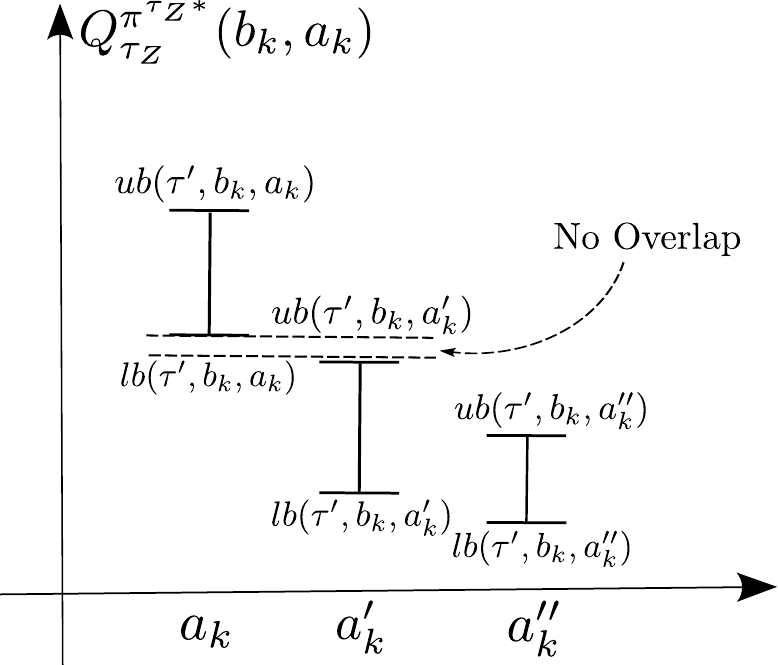}
		\caption{No overlap for topology $\topo'$}
		\label{fig:guarantee_b}
	\end{subfigure}
	\caption{(a) Bounds over the optimal Q function considering topology $\topo$. Due to the indicated overlap, the optimal action cannot be deduced. (b) The bounds can be tightened by switching to a different topology $\topo'$, as described in the text, until \eqref{eq:sameaction} is satisfied,  and the optimal action can be determined.
} 
	\label{fig:guarantee} 
	\vspace{-5pt}
\end{figure}

Given such bounds it is possible to identify the optimal action of the original POMDP,  $\action^{\star}_k\triangleq \argmax_{\action_k \in \CA} Q^{\policy^{\topo_Z*}}_{\topo_Z}(b_k,\action_k)$, when 
\begin{equation}
\exists \bar{a}_k\in \CA, \text{s.t.}  \ lb(\topo, \policy^\topo, b_k, \bar{a}_k) > ub(\topo, \policy^\topo, b_k, a_k) \ \forall a_k \in \CA\setminus \{\bar{a}_k\},
\label{eq:sameaction}
\end{equation}
and assigning $a^{\star}_k=\bar{a}_k$. Such a situation is illustrated in Fig.~\ref{fig:guarantee_b}. This achieves a formal performance guarantee, getting the same optimal action $a_k^*$ as the original full POMDP. Moreover, if the bound $B(\topo, \policy^{\topo}, b_k, a_k)$ does not depend on the original full observation space and model, we can avoid building the original belief tree $\BelTree^{\topo_Z}$. 

In case the condition \eqref{eq:sameaction} is not satisfied, as illustrated in Fig.~\ref{fig:guarantee_a}, we can no longer guarantee the optimal action $a_k^{\star}$ will be selected. In such a case, there are several options: (i) determine the action using either the optimal simplified Q-function, i.e.~$\argmax_{\action_k \in \CA} Q^{\policy^{\topo*}}_{\topo}(b_k,\action_k)$, or using the bounds, e.g.~the action with the highest lower  or upper bound, while bounding the worst-case loss in planning performance (regret), similar to e.g.~\cite{Elimelech22ijrr, Zhitnikov24ijrr}; (ii) tighten the bounds until the condition \eqref{eq:sameaction} is met.  The latter can be done either by considering different policies in a given topology $\topo$, or by switching to another topology. 

\vspace{+5pt}

The tightest bounds for a given topology can be obtained as follows,
\begin{align}
	lb(\topo, b_k, a_k)  &= \max_{\policy^\topo \in \Pi^{\topo}} lb(\topo, \policy^\topo, b_k, a_k) \label{eq:LBoundOptQgeneralTightest} 
	\\
	ub(\topo, b_k, a_k)  &= \min_{\policy^\topo \in \Pi^{\topo}} ub(\topo, \policy^\topo, b_k, a_k).
	\label{eq:UBoundOptQgeneralTightest}
\end{align}
In general, these bounds are \emph{not} necessarily obtained for the optimal policy $\policy^{\topo*}=\argmax_{\policy^\topo \in \Pi^{\topo}} V^{\policy^\topo}_{\topo}(b_k)\equiv \argmax_{\policy^\topo \in \Pi^{\topo}} Q^{\policy^\topo}_{\topo}(b_k, \policy^\topo_k(b_k))$. Moreover, it is possible that even with these tightest bounds, the condition \eqref{eq:sameaction} is not satisfied, in which case we have to switch to another topology, as discussed in Section \ref{sec:guarantee_for_case}.

We provide a conceptual illustration of the above in Fig.~\ref{fig:guarantee}, considering three possible actions, $a_k, a'_k$ and $a''_k$. In Fig. \ref{fig:guarantee_a}, there is no overlap between the bounds of $a_k$ and $a_k''$. We can conclude that the action $a_k$ is definitely better than $a_k''$, and can safely prune it. However, we cannot distinguish between $a_k$ and $a_k'$ due to the overlap between the bounds. In order to find the optimal action, we have to try another simplified topology $\topo'$ as in Fig. \ref{fig:guarantee_b}. With the new topology $\topo'$, there is no bound overlap for action $a_k$ with respect to other actions, i.e.~the condition \eqref{eq:sameaction} is met, indicating $a_k^{\star}=a_k$ is the best action. Thus, we find the optimal action $a_k^{\star}$ for the original POMDP without exploring the original complicated belief tree $\BelTree^{\topo_Z}$, and only exploring different simplified belief trees.


For the general setting considered herein, we now provide one possible bound $B(\topo, \policy^{\topo}, b_k,a_k)$ from \eqref{eq:Boundgeneral} that is valid without any further assumptions. Specifically, we propose to use the QMDP as the upper bound of the POMDP~\cite{Ross08jair}, 
\begin{flalign}
	B(\topo, \policy^{\topo}, b_k,a_k) = \max_{\policy^{QMDP}} \big| Q^{\policy^\topo}_{\topo}(b_k,\action_k)-Q^{\policy^{QMDP}}(b_k,\action_k)\big|.
\end{flalign}
To retrieve this bound, we only need to explore a smaller policy space without any observations (see connection of our approach to QMDP in Remark \ref{remark1}). 

It is worth noting that this is a general bound that is valid for all definitions of alternative observation space and model. However, the bound is dependent on the specific choice of the alternative observation space and model. A specific choice of alternative observation space and model can lead to a better bound, as discussed next.

\section{Specific Case: Full Observability}
\label{subsec:full-obs-def}
In this section, we consider a specific instantiation of our framework from Section \ref{sec:AdatpiveObservationTree}, where the state is fully observable.  The corresponding alternative observation space $\mathcal{O}$ and model $\probo{o\mid x}$ are therefore defined as,
\begin{align}
	\probo{o \mid x} \triangleq \delta(o-x), \ 
	\ \text{where }\
	o \in \Ospace \triangleq \CX.
	\label{equ:dirac-obs-model}
\end{align}
%
%
%
%
%
As seen, the alternative observation space is set to be the state space, and $\probo{o\mid x}$ is a Dirac function.
 For the topology $\topo$, if a single belief node $b^{\topo-}_i$ switches to the alternative observation space and model, this will reduce the dimension of the policy space for $b^{\topo-}_i$, $\{\policy_i(b^{\topo-}_i, \gobs_i)\}$ from
$|\mathcal{Z}||\mathcal{A}|$ to $|\mathcal{X}||\mathcal{A}|$. This reduces the number of child posterior belief nodes if $|\mathcal{X}|<|\mathcal{Z}|$.


For the original full belief tree, we now consider calculation of the expected state-dependent reward at any depth $i+1$, $\mathbb{E}_{x_i|b^{\topo-}_i}\mathbb{E}_{\bar{z}_i|x_i, h^{\topo-}_i}\mathbb{E}_{x_{i+1}|x_i, a_i}[r(x_{i+1})]$]. The corresponding complexity for the original observation space is thus $|\mathcal{Z}||\mathcal{A}||\mathcal{X}|^2$. In contrast, for the alternative observation space and model \eqref{equ:dirac-obs-model}, the complexity becomes $|\mathcal{A}||\mathcal{X}|^2$. 





\subsection{Performance Guarantees}\label{subsuc:bound-full-obs}

Having defined the specific alternative observation space and model \eqref{equ:dirac-obs-model}, we are now interested in providing formal planning performance guarantees by bounding the Q function of the original POMDP, that corresponds to topology $\topo_{Z}$,  and the simplified POMDP considering some topology $\topo$. Specifically, considering, at this point, some arbitrary policies $\policy^{\topo_Z}$ and $\policy^{\topo}$ for the two topologies, we aim to bound 
\begin{equation}
	\Delta Q(b_k,\action_k, \policy^{\topo_Z},\policy^{\topo},\topo_Z, \topo) \triangleq \big| Q^{\policy^\topo}_{\topo}(b_k,\action_k)-Q^{\policy^{\topo_Z}}_{\topo_Z}(b_k,\action_k) \big|.
	\label{eq:DiffQ}
\end{equation}
Note that for the optimal policy $\policy^{\topo *}$ we get back to \eqref{eq:Boundgeneral}.

We start with bounding the difference between the expected immediate reward of two different topologies $\topo$ and $\topo'$, where $\topo'$ has fewer belief nodes using an alternative observation space. In particular, $\topo'$ could correspond to $\policy^{\topo_Z}$. 
All the proofs for this section can be found in the Supplementary document \cite{Kong24isrr_supplementary}.

\begin{lemma}
	\label{lemma:bound-r}
	Consider two topologies $\topo$ and $\topo'$, where $\topo'$ has fewer belief nodes using the alternative observation space.  The difference between the expected state-dependent rewards at any time instant $i$,  considering policy $\policy^{\topo'{}}_i$ for topology $\topo'$ and policy $\policy^{\topo}_{i}$ for topology $\topo$ is bounded as: 
	\begin{flalign}
			&\Big|\exptsimp{\gobs_{1:i}|b_k, \pi^{\topo}}{\topo}(r(b^{\topo}_{i})) -  \exptsimp{\gobs_{1:i}|b_k, \pi^{\topo'}}{\topo'}(r(b^{\topo'}_{i})) \Big|
			\\ &\leq \max_{\bpolicy^{\topo}\in \Pi^{\topo}}\Big|\exptsimp{\gobs_{1:i-1}|b_k, \pi^{\topo}}{\topo} \emathbb{x_i|h^{\topo-}_i}r(x_i) \nonumber \\ &-\emathbb{x_0|b_k}\emathbb{x_1|x_0,\bpolicy^{\topo}_0}\exptsimp{\gobs_1|x_1,h^{\topo-}_{1}}{\topo}...\emathbb{x_{i-1}|x_{i-2},\bpolicy^{\topo}_{i-2}} \exptsimp{\gobs_{i-1}|x_{i-1},h^{\topo-}_{i}}{\topo}\emathbb{x_i|x_{i-1},\bpolicy^{\topo}_{i-1}}r(x_i) \Big|.
	\end{flalign}
\end{lemma}
We can use a similar method to also bound the difference between the Q functions of different topologies.
\begin{lemma}
	The difference \eqref{eq:DiffQ} between the Q functions of the original and simplified POMDPs, represented by topologies $\topo'$ and $\topo$, can be bounded by exploring the simplified policy space $\Pi^{\topo}$ as:
		\begin{equation}
			\Delta Q(b_k,\action_k, \policy^{\topo'},\policy^{\topo},\topo', \topo)
			\leq 
			\max_{\bpolicy^{\topo}\in \Pi^{\topo}} | Q_{\topo}^{{\bpolicy^{\topo}}}(b_k,\action_k) - Q_{\topo}^{\policy^{\topo}}(b_k,\action_k) | 
			\triangleq
			\delta_Q(b_k,\action_k, \policy^{\topo},\topo),
		\nonumber
	\end{equation}
	where $\policy^{\topo'}\in \Pi^{\topo'}$ and $\pi^{\topo}\in \Pi^{\topo}$ are some policies in the original and simplified policy spaces, $\Pi^{\topo'}$ and $\Pi^{\topo}$, respectively.
   \label{lemma:bound-q}
\end{lemma}

\begin{theorem}
\label{theorem:bound-q}
Consider two topologies $\topo$ and $\topo'$, where $\topo'$ uses fewer belief nodes with the alternative observation space. Then, 
	we can bound the Q function of topology $\topo'$, by deriving the tightest bound from Lemma \ref{lemma:bound-q}: 
		\begin{align}
		\min_{\policy^{\topo}\in \Pi^{\topo}}[Q_{\topo}^{\policy^{\topo}}(b_k,\action_k)] \leq 
		Q_{\topo'}^{\policy^{\topo'}}(b_k,\action_k) \leq \max_{\policy^{\topo}\in \Pi^{\topo}}[Q_{\topo}^{\policy^{\topo}}(b_k,\action_k)].
		\label{eq:BoundQ-any}
	\end{align}
	Specifically, $\topo'$ can be the full topology $\topo_Z$, which represents the original POMDP:
	\begin{equation}
		 \min_{\policy^{\topo}\in \Pi^{\topo}}[Q_{\topo}^{\policy^{\topo}}(b_k,\action_k)] \leq 
	Q_{\topo_Z}^{\policy^{\topo_Z}}(b_k,\action_k) \leq \max_{\policy^{\topo}\in \Pi^{\topo}}[Q_{\topo}^{\policy^{\topo}}(b_k,\action_k)].
			\label{eq:BoundQ}
	\end{equation}
\end{theorem}
%
%
Since the bounds \eqref{eq:BoundQ} are valid for any policy $\pi_Z\in \Pi_Z$, we can utilize them to bound the optimal Q function for a corresponding optimal policy $\policy^{\topo_Z*}$, i.e.~
\begin{align}
	lb(\topo, b_k, \action_k) \leq 
	Q_{\topo_Z}^{\policy^{\topo_Z*}}(b_k,\action_k) \leq  ub(\topo, b_k, \action_k),
	\label{eq:BoundsFullObserv}
\end{align}
where
\begin{align}
	ub(\topo, b_k, \action_k) \triangleq \max_{\policy^{\topo}\in \Pi^{\topo}}[Q_{\topo}^{\policy^{\topo}}(b_k,\action_k)], \ \ 
	lb(\topo, b_k, \action_k) \triangleq 
	\min_{\policy^{\topo}\in \Pi^{\topo}}[Q_{\topo}^{\policy^{\topo}}(b_k,\action_k)].
	\label{eq:upperboundQ}
\end{align}
Moreover, we can get a tighter lower bound for $Q_{\topo_Z}^{\policy^{\topo_Z*}}(b_k,\action_k)$ of the optimal policy in an iterative process as follows. 
\begin{theorem}
	\begin{align}
		Q_{\topo_Z}^{\policy^{\topo_Z*}}(b_k,\action_k) \geq lb(b_k, a_k, \topo),		
	\end{align}	
	where $lb(b_k, a_k, \topo)$ is defined recursively for $t\in [k+1,k+L-1]$ as
	\begin{align}
		\label{equ:lq-def}
		lb(b_t^{\topo}, &a_t, \topo) \triangleq \simpfunc^{\topo}(\shis_{t+1}^{\topo-})[r(b_{t}^{\topo},\action_t) +\exptsimp{\gobs_{t+1}|\shis^{\topo}(b_{t}^{\topo}),\action_t}{\topo} \max_{\policy_{t+1}^{\topo}}lb(b_{t+1}^{\topo},\policy_{t+1}^{\topo}(b_{t+1}^{\topo}),\topo)] \\
		+& (1-\simpfunc^{\topo}(\shis_{t+1}^{\topo-}))[r(b_{t}^{\topo},\action_t) +\exptsimp{\gobs_{t+1}|\shis^{\topo}(b_{t}^{\topo}),\action_t}{\topo} \min_{\policy_{t+1}^{\topo}}lb(b_{t+1}^{\topo},\policy_{t+1}^{\topo}(b_{t+1}^{\topo}),\topo)], \nonumber
	\end{align} 
	where $\shis_{t+1}^{\topo-} = \{ \shis_{t}^{\topo}(b_t^{\topo}), a_t \}$  	and $	lb(b_L^{\topo}, a_L, \topo)= r(b^{\topo}_L,a_L)$.
	\label{theorem:bound-optimal-q}
\end{theorem}	

Overall, we obtained upper and lower bounds for the optimal Q function in the original POMDP by only exploring a simplified POMDP represented by topology $\topo$. 
These bounds are specific to the considered alternative space and model \eqref{equ:dirac-obs-model}, as opposed to the previously shown general bounds \eqref{eq:LBoundOptQgeneralTightest} and \eqref{eq:UBoundOptQgeneralTightest}. Thus we only need to explore a subset of the policy space of the original POMDP.

\begin{figure}[t]
	\centering
	\includegraphics[width=0.7\textwidth]{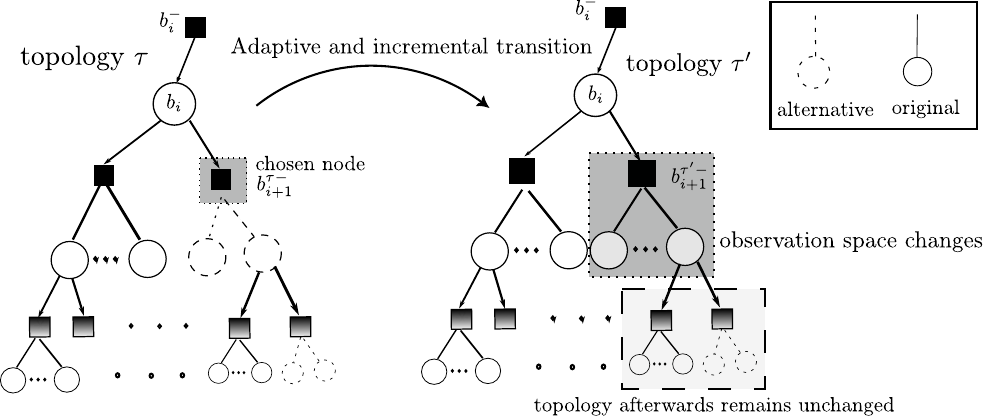}
	\vspace{-5pt}
	\caption{A conceptual illustration of  incremental and adaptive transition from topology $\topo$ to $\topo'$.}
	\label{fig:cross-topo}
\end{figure}


\begin{remark}[Connections with QMDP]\label{remark1}
Our simplification of POMDP by an alternative observation space and model has a connection to QMDP planning. 
If the topology is chosen to be $\topo_O$, where all the belief nodes have the alternative observation space and model \eqref{equ:dirac-obs-model}, our adaptive belief tree will become the QMDP belief tree. Hauser \cite{Hauser10wafr} exploited QMDP approximation to gather information during planning. QMDP bound is also used as an upper bound for POMDP~\cite{Ross08jair}. However, in practice, the pure QMDP approximation usually has a loose bound, which cannot be used to identify the optimal action, i.e.~the condition \eqref{eq:sameaction} is not met. In stark contrast, our method  uses  an adaptive structure to switch observation space and model at only parts of the belief nodes while providing formal guarantees in terms of \eqref{eq:BoundsFullObserv} and Theorem \ref{theorem:bound-optimal-q}. 
It is noted that our method actually reveals the underlying principle of the QDMP approximation, which can be considered as a special case within our scheme. 

\end{remark}
\vspace{-10pt}

\subsection{Bounds Analysis}\label{sec:guarantee_for_case}

\textbf{Convergence.}
If we keep transitioning between topologies, at each iteration turning more nodes back to the original observation space and model, the final topology will be $\topo_Z$. 
The upper bound of $\topo_Z$ will be the optimal Q function:
$ub(\topo_z, b_k, a_k) =  \max_{\policy^{\topo_Z}}[Q_{\topo_Z}^{\policy^{\topo_Z}}(b_k,\action_k)] = Q_{\topo_Z}^{*}(b_k,\action_k)$. Similarly, the corresponding lower bound will also become the optimal Q function, $lb(\topo_z, b_k, a_k) =  Q_{\topo_Z}^{*}(b_k,\action_k)$, by iterating Theorem \ref{theorem:bound-optimal-q}.

\textbf{Monotonicity.}
Here, we consider the same setting of the topology $\topo$ and a less simplified topology $\topo'$ same as the Section \ref{subsuc:bound-full-obs}. 
We can use the same technique as Theorem \ref{theorem:bound-q} to prove that the upper bound will be tightened:
\begin{equation}
	uq(\topo, b_k, a_k) = \max_{\policy^{\topo}}[Q_{\topo}^{\policy^{\topo}}(b_k,\action_k)] \geq \max_{\policy^{\topo'}}[Q_{\topo'}^{\policy^{\topo'}}(b_k,\action_k)] = uq(\topo', b_k, a_k).
\end{equation}
The lower bound will also be tightened.  Let us assume that topology $\topo'$ turns only  a single belief node, that had an alternative observation space and model in $\topo$, back to the original observation space and model. Without losing generality, assume this node is located at some depth $i+1$. Therefore, all the belief nodes in $\BelTree^{\topo}$ and $\BelTree^{\topo'}$ at depth $i$ (or smaller) are identical,  
$b^{\topo}_i = b^{\topo'}_i\triangleq b_i$. Then we have:
\begin{flalign}
	lb(\topo', b_i, a_i) &= r(b_i,a_i) +\emathbb{\gobs_{i+1}} \max_{\policy^{\topo'}_{i+1}}lb(b^{\topo'}_{i+1},\policy^{\topo'}_{i+1},\topo') \\ &\geq r(b_i,a_i) +\emathbb{\gobs_{i+1}} \min_{\policy^{\topo}_{i+1}}lb(b^{\topo}_{i+1},\policy^{\topo}_{i+1},\topo) = lb(\topo, b_i, a_i).
\end{flalign}
If we keep iterating the inequality back to the root, we will see the lower bound will be tightened from $\topo$ to $\topo'$:
$
	lb(\topo', b_k, a_k) \geq lb(\topo, b_k, a_k).
$
This process can be generalized to show that the bound becomes tighter also considering the topology $\topo'$ turns a number of nodes back to the original observation space and model.

\textbf{Incremental transition between topologies.}
We now briefly describe the calculations involved in transitioning from topology $\topo$ to $\topo'$. Assume in the latter, some number $n$ of propagated history (belief) nodes from $\topo$ that had an alternative observation space are switched to the original observation space. Let $\tilde{H}$ be the set of these nodes. When transitioning from $\topo$ to $\topo'$, the Q function of some belief nodes $h^r$ will not change, and we can reuse them without recalculation.  
These nodes are located in branches that are common in $\topo$ and $\topo'$, i.e.~there does \emph{not} exist an ancestor or descendant belief node that is included in $\tilde{H}$. 

Moreover, consider histories in $H^r$ that correspond to top-level (minimum depth) beliefs in $\topo'$ with respect to all the beliefs represented by $H^r$, i.e.~for each such history $h_{\ell}\in H^r$ of some depth $\ell$ and the corresponding belief $b_{\ell}$, there does not exist another history $h_{\ell'}\in H^r$ with $\ell'<\ell$ and a corresponding belief $b_{\ell'}$ that is an ancestor of $b_{\ell}$. By definition, all beliefs in $\topo'$ that are ancestors of these minimum-depth beliefs are identical in the two topologies $\topo$ and $\topo'$. Therefore, as we traverse the tree from the leaves upwards and identify the optimal action in each branch by maximizing the Q function, at some point, we will reach 
one of the minimum-depth beliefs in $H^r$. At this point, while proceeding upwards the tree, if we identify that the optimal value functions of \emph{all} the $b_{l'}^{\topo'}$ at a level $l'<l$ do not change with respect to $\topo$, this means the optimal Q function in that level is determined by a branch that is not affected by the switch from $\topo$ to $\topo'$. Based on this, it is \emph{no longer} necessary to keep calculating upwards, which will lead to a further speedup in planning. See illustration in Fig.~\ref{fig:cross-topo}.
\vspace{-10pt}

\section{Estimator}\label{sec:Estimator}
\vspace{-5pt}
While thus far, we considered exact calculations of the Q function and of the bounds, in practice, this is only possible for small problems and is limited to discrete spaces. In larger and more realistic problem settings, we need to consider a POMDP solver that constructs an estimator of the (optimal) Q functions. In this section, we propose to use the sparse sampling method \cite{Kearns02ml}, to estimate the upper and lower bounds derived in Section \ref{subsec:full-obs-def}, considering the specific alternative observation model and space \eqref{equ:dirac-obs-model}.

\begin{wrapfigure}{L}{0.4\textwidth}
	\centering
	\includegraphics[width=\textwidth]{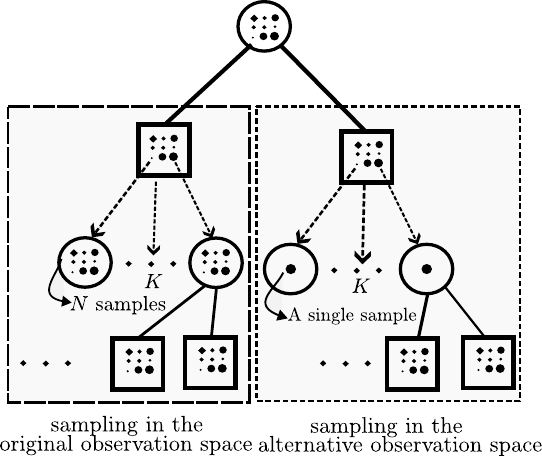}
	\caption{A sparse sampling belief tree using the original and alternative observation space and models \eqref{equ:dirac-obs-model}. \vspace{-10pt}}
	\label{fig:sample}
\end{wrapfigure}
Specifically, we consider a particle-based belief representation $\{x^i, w^i\}_{i=1}^N$, where $N$ is the number of particles and $w^i$ are the unnormalized weights. The corresponding particle belief is then
 $b(x)\triangleq \frac{\sum_{i=1}^N w^i \delta(x-x^i)}{\sum_{j=1}^N w^j}$. For our simplified belief tree with the adaptive structure, the sampling process is a bit different from the common method. Fig.~\ref{fig:sample} illustrates this process. For the propagated belief nodes that use an alternative observation space and model \eqref{equ:dirac-obs-model}, we only generate a single sample for the new posterior belief node due to the deterministic observation model in Equation (\ref{equ:dirac-obs-model}). This is in contrast to generating $N$ samples for posterior belief nodes using the original observation space. After this step, we generate $N$ samples that represent the propagated belief; these particles are sampled from the transition model given the single sample from the previous posterior belief and the corresponding action (see Fig.~\ref{fig:sample}). Compared to the sparse sampling in the original belief tree, our simplification can reduce the calculation complexity of the state-dependent reward at each belief node from $O(N)$ to $O(1)$. This reduction is significant for a large  $N$.

\textbf{Bounded Estimation Error.}
We define the estimated upper and lower bound as $\hat{ub}(b_0,a_0, \topo)$ and $\hat{lb}(b_0,a_0, \topo)$. 
The estimation of the upper bound can be seen as an estimation of the optimal Q function within the belief tree $\BelTree^\topo$:
\begin{equation}
	\hat{ub}(b_0,a_0, \topo) = 
	\max_{\policy^{\topo}_{1:L-1}}	[\hat{Q}_{\topo}^{\policy^{\topo}_{1:L-1}}(b_0,\action_0)] 
	=  \hat{Q}_{\topo}^{\policy^{\topo}*}(b_0,\action_0).
\end{equation}
The estimation of the lower bound $\hat{lb}(b_0,a_0, \topo)$ should be done iteratively following the recursive form  Theorem \ref{theorem:bound-optimal-q}. 
Then, we define the estimation error as
\begin{align}
	\Delta \hat{ub}(b_0,a_0, \topo) &\triangleq |\hat{ub}(b_0,a_0, \topo) - {ub}(b_0,a_0, \topo)|, 
	\\
	\Delta \hat{lb}(b_0,a_0, \topo) &\triangleq|\hat{lb}(b_0,a_0, \topo) - {lb}(b_0,a_0, \topo)|, 
\end{align}
where for the upper bound, $\Delta \hat{ub}(b_0,a_0, \topo) = |\hat{Q}_{\topo}^{\policy^{\topo*}}(b_0,\action_0) - {Q}_{\topo}^{\policy^{\topo*}}(b_0,\action_0) |$.
%

With the estimator bounds, we can use the estimated upper and lower bounds to determine the optimal action in case \eqref{eq:sameaction} is satisfied. If not, we will follow the same procedure to switch between topologies as introduced in Section \ref{subsec:performance-guarantee}. 
We now bound probabilistically the estimation error of these bounds, utilizing the Hoeffding inequality and similar derivations to \cite{Lim23jair, Kearns02ml}. 

\begin{theorem}
	[\textbf{Bounded Estimation Error}] For all the depth $d=0,\ldots ,L-1$ and $a_d$, the following concentration bound holds with probability at least
	$1-2|A|(|A|C)^{L-d}\exp(\frac{-C\lambda^2}{2V_{\max}^2})$ :
		\begin{equation*}
		\Delta \hat{ub}(b_d,a_d, \topo) \leq \frac{(L-d)(L-d-1)}{2}\lambda,
		\Delta \hat{lb}(b_d,a_d, \topo) \leq \frac{(L-d)(L-d-1)}{2}\lambda.
	\end{equation*}
Specifically, for $d=0$, we obtain probabilistic bounds on the estimation error of  $\hat{ub}(b_0,a_0, \topo)$ and $\hat{lb}(b_0,a_0, \topo)$ at the root of the belief tree.
	\label{theorem:upper-bound-estimation}
\end{theorem}
%
This Theorem provides guarantees that the sparse sampling method can estimate our proposed upper and lower bound well with a probabilistically bounded error. 

%
\vspace{-10pt}

\section{Experiments}
We evaluate our proposed method through POMDP simulations conducted in three distinct settings. Firstly, we investigate the exact computation of the original POMDP and the proposed bounds to validate the findings of the theoretical analysis presented in Section \ref{subsec:full-obs-def}. Secondly, we assess a sparse sampling POMDP solver using our estimated bounds, as outlined in Section \ref{sec:Estimator}, addressing larger POMDP problems. Finally, we apply the sparse sampling method to a beacon navigation problem to showcase the potential of our approach in practical robotics applications. The detailed experiment settings appear in Section \ref{sec:experiment-settings} of the Supplementary document  \cite{Kong24isrr_supplementary}.


\vspace{-6pt}
\subsection{Exact Full Calculation}
\begin{wrapfigure}{l}{0.4\textwidth}
	\centering
	\includegraphics[width=\textwidth]{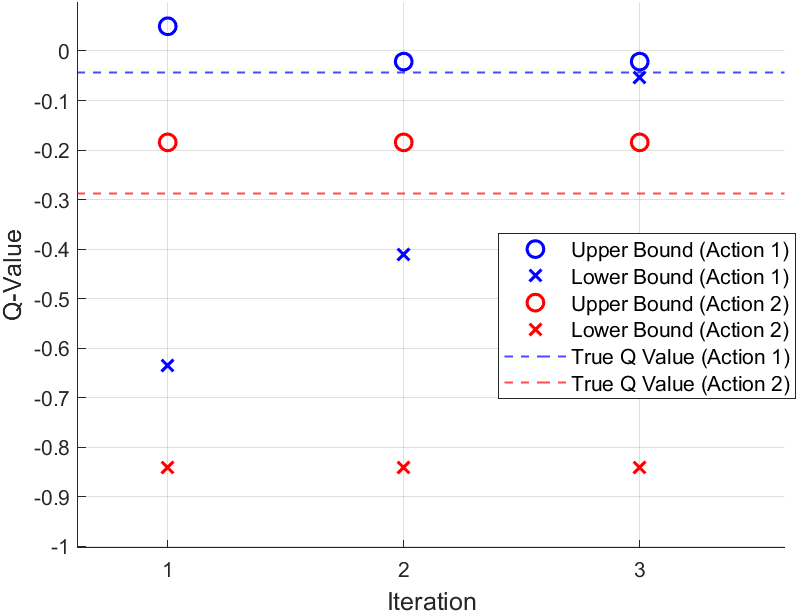}
	\caption{Bounds over $Q^{\topo_Z*}(b_k,a_k)$ considering exact calculations. 
	}
	\label{fig:result1}
\end{wrapfigure}

We start with exactly calculating the original Q function at the root $Q^{\topo_Z*}(b_k,a_k)$ and the upper and lower bound, \eqref{eq:upperboundQ} and \eqref{equ:lq-def}. To that end, we consider a small POMDP problem for which an exact solution can be calculated in a reasonable time. Specifically, we utilize the random POMDP library to generate a discrete POMDP problem with observation, action, and state spaces of $|\mathcal{Z}|=20$,  $|\mathcal{A}|=2$, and $|\mathcal{X}|=3$, respectively and set the planning horizon to $L=2$. 
Here, we set the observation space to be much larger than the state space in order to represent the kind of POMDP where the observation space is large.

\begin{wraptable}{l}{0.33\textwidth}
	\centering
	\caption{Comparison of methods for an exact calculation of the Q function. 
	}
	\label{tab:result1}
	\resizebox{\linewidth}{!}{
		\begin{tabular}{lcc}
			\hline
			Method &  Total Cost Time (s) \\
			\hline
			Proposed  & \textbf{0.965} \\
			Full Problem  & 2.675 \\
			\hline
		\end{tabular}
	}
	\vspace{-5pt}
\end{wraptable}
We show the evolution of the upper $ub(\topo, b_k, \action_k)$ and lower $lb(\topo, b_k, \action_k)$ bounds on $Q^{\topo_Z*}(b_k,a_k)$ during this  process in Fig.~\ref{fig:result1}, where each iteration corresponds to a different topology $\topo$.  We report planning time in  Table \ref{tab:result1}. As seen, after 3 iterations (topology switches) our method  finds the optimal action $a_k^*= \text{Action 1}$, with the total planning time of $0.965$s versus $2.675$s that corresponds to solving the original POMDP exactly in this toy example.

\vspace{-6pt}
\subsection{Online Estimator}
\begin{wraptable}{l}{0.33\textwidth}
	\centering
	\caption{Comparison of Sparse Sampling for a Discrete POMDP. 
		For $N=50$ and $N=90$, our method explores $3$ and $2$ different topologies, respectively. 
		%
		}
	\label{tab:result2}
	\resizebox{\linewidth}{!}{
\begin{tabular}{@{}lcc@{}} 
\toprule
\multicolumn{1}{c}{\multirow{2}{*}{Method}} & \multicolumn{2}{c}{Running Time (s)} \\ \cmidrule(l){2-3}
\multicolumn{1}{c}{}                      & \multicolumn{1}{l}{$N=50$} & \multicolumn{1}{l}{$N=90$} \\ \midrule
Our Simplification                      & \textbf{0.869}             & \textbf{0.880}             \\
Original POMDP                           & 1.782             & 3.276             \\
\bottomrule
\end{tabular}
	}
\end{wraptable}

Here, we use a sparse sampling estimator to estimate the upper and lower bounds practically.
We generate a discrete POMDP problem with observation, action, and state spaces $|Z|=2000$, $|A|=2$, and $|X|=1000$, respectively.


The results of the experiment are reported in Table \ref{tab:result2}. A comparison was conducted between two distinct sets of sampling parameters: $K=N=90$ versus $K=N=50$. With the larger values of $K$ and $N$, our method demonstrated the capability to identify the optimal action with reduced iterations and a significant decrease in planning time. Notably, in both scenarios, our method successfully identified the same optimal action as the sparse sampling employed in the original POMDP tree.


\vspace{-10pt}
\subsection{Beacon Navigation Problem}

\begin{figure}
    \centering
    \begin{subfigure}{0.35\textwidth}
        \includegraphics[width=\textwidth]{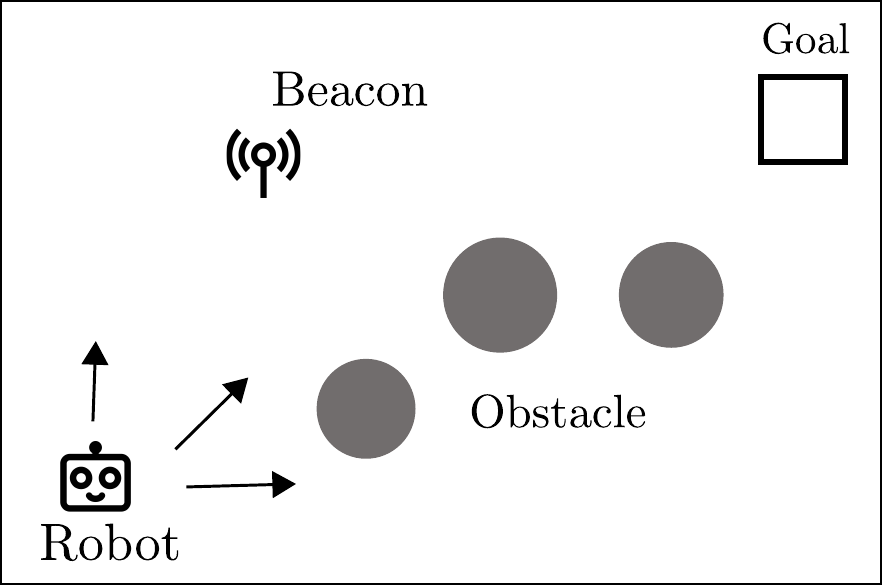}
        \caption{
        }
        \label{fig:result3-a}
    \end{subfigure}
    \hfill
    \begin{subfigure}{0.3\textwidth}
        \includegraphics[width=\textwidth]{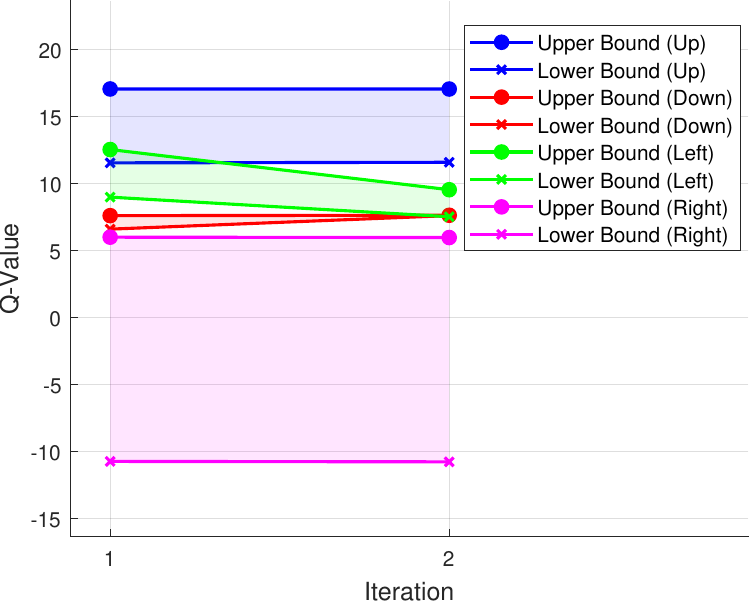}
        \caption{
        }
        \label{fig:result3-b}
    \end{subfigure}
    \hfill 
    \begin{subfigure}{0.25\textwidth} 
        \centering
	\resizebox{\linewidth}{!}{
		\begin{tabular}{lcc}
			\hline
			Method  & Planning Time (s) \\
			\hline
			Proposed   & \textbf{1.046} \\
			Original &  2.969 \\
			\hline
		\end{tabular}
	}
        \caption{
        }        
        \label{tab:beacon-result}

    \end{subfigure}

    \caption{Beacon Navigation Problem. (a) Scenario. (b) Bounds over the estimated optimal Q function as a function of iteration number. Our method iterates two different topologies. (c) Planning time. }
    \label{fig:result3}
\end{figure}

In this experiment, we utilize the sparse sampling estimator to address a specific problem in robot navigation, referred to as the beacon navigation problem.
The objective is for the robot to navigate a 2D space, maneuvering around obstacles to reach a specified goal, while localizing itself based on observations from known beacons. The scenario is shown in Fig.~\ref{fig:result3-a}.  
In this setting, the POMDP problem has a continuous state and observation space and a discrete action space of $|A|=4$. We provide further details regarding this scenario in the Supplementary \cite{Kong24isrr_supplementary}. 
Table \ref{tab:beacon-result} presents timing results for the first planning session. During each iteration, the considered topology switches back $5$ nodes to use the original observation space and model. 
Notably, our proposed approach switches twice between different topologies to identify the same optimal action at the outset, demonstrating superior efficiency compared to a conventional sparse sampling in the original belief tree. Fig.~\ref{fig:result3-b}  shows the corresponding bounds over the optimal Q-function at the root of the belief tree in this process. 



\vspace{-10pt}

\section{Conculsion}

We proposed a novel framework to speedup POMDP planning by selectively switching part of the belief nodes to an alternative observation space and model while providing formal performance guarantees with respect to the original POMDP problem. We defined the notion of adaptive topology belief trees 
 and examined a specific definition of the alternative observation space that corresponds to a fully observable setting. In such a setting, we derived novel bounds that can be used to adaptively switch between different topologies until the optimal action of the original problem can be determined. 
 We also demonstrated a bound across different policy spaces induced by different belief tree topologies, indicating a new way to simplify the policy space. The experiments support our claim, leading to a significant speedup in planning (e.g.~$\times 3$) while identifying the same action as with the original observation space. We believe this work opens new exciting avenues for online POMDP planning with formal performance guarantees.

\vspace{-10pt}

\subsection*{Acknowledgment}
	This work was supported by the Israel Science Foundation (ISF). 
	 Da Kong was partially supported by the Lady Davis Scholarship.

%
%
\bibliographystyle{IEEEtran}
\bibliography{refs}

\newpage
\appendix

\section{Proof of Lemma \ref{lemma:bound-r}}\label{sec:Proof_lemma:bound-r}
\subsection{Preliminary proof}
\label{pre-proof}
	For a given transition model $\prob{x_{i+1}|x_i, a_i}$ and a given observation model $\prob{z_i| x_i}$, if we assume the history $h_i^-$ and $h_i^{\topo-}$ are known, we can bound the expected state-dependent reward below between the original policy $\policy^{\topo{Z}}_i$ and a known policy $\policy^{\topo}_{i}$ for topology $\topo$ as: 
\begin{align}
	&\big|\emathbb{x_i|h^-_i}\exptsimp{\gobs_i|x_i, h^-_i}{\topo_Z}\emathbb{x_{i+1}|x_i, \policy^{\topo{Z}}_i}r(x_{i+1}) - \emathbb{x_i|h^{\topo-}_i}\exptsimp{\gobs_i|x_i, h^{\topo-}_i}{\topo}\emathbb{x_{i+1}|x_i, \policy^{\topo}_i}r(x_{i+1})\big| \\
	&\leq \max_{\bpolicy_i} \Big|\emathbb{x_i|h^-_i}\exptsimp{\gobs_i|x_i, h^-_i}{\topo}\int_{x_{i+1}}\prob{x_{i+1}|x_i, \bpolicy_i(h^-_i,\gobs^{\topo}_i)}r(x_{i+1}) \\&- \emathbb{x_i|h^{\topo-}_i} \exptsimp{\gobs_i|x_i, h^{\topo-}_i}{\topo}\int_{x_{i+1}}\prob{x_{i+1}|x_i, \policy^{\topo}_i}r(x_{i+1})\Big| 
\end{align}
Here, the policy $\bpolicy_i$ is a mapping : $\mathcal{H}^-_i \times \gobSpace_i(\mathcal{H}^-_i, \topo) \mapsto \CA.$

 \begin{proof}
	
	\textbf{First, we consider the situation that the belief node uses the alternative observation space and model: $\beta^{\topo}(h^-_i)=0$.}
	
	For each $x_i \in \CX$, we can find an action $a_i$ that maximize and minimize the expected reward:
	\begin{equation}
		\int_{x_{i+1}}\prob{x_{i+1}|x_i, \policy^{\topo_Z}_i} r(x_{i+1}) dx_{i+1} \leq  \max_{a_i(x_i) \in \CA}\int_{x_{i+1}}\prob{x_{i+1}|x_i, a_i} r(x_{i+1}) dx_{i+1}
		\label{eq-14-proof1}
	\end{equation}
	\begin{equation}
		\int_{x_{i+1}}\prob{x_{i+1}|x_i, \policy^{\topo_Z}_i} r(x_{i+1}) dx_{i+1} \geq  \min_{a_i(x_i) \in \CA}\int_{x_{i+1}}\prob{x_{i+1}|x_i, a_i} r(x_{i+1}) dx_{i+1}
		\label{eq-15-proof1}
	\end{equation}
	From (\ref{eq-14-proof1}), we have:
	\begin{align}
		& \nonumber \emathbb{x_i|h^-_i}\exptsimp{\gobs_i|x_i, h^-_i}{\topo_Z}\int_{x_{i+1}}\prob{x_{i+1}|x_i, \policy^{\topo_Z}_i} r(x_{i+1}) dx_{i+1} \\ &\leq  \emathbb{x_i|h^-_i}\exptsimp{\gobs_i|x_i, h^-_i}{\topo_Z}\max_{a_i(x_i) \in \CA}\int_{x_{i+1}}\prob{x_{i+1}|x_i, a_i} r(x_{i+1}) dx_{i+1} \\
		& = \emathbb{x_i|h^-_i}\exptsimp{\gobs_i|x_i, h^-_i}{\topo}\max_{a_i(x_i) \in \CA}\int_{x_{i+1}}\prob{x_{i+1}|x_i, a_i} r(x_{i+1}) dx_{i+1}
		\label{eq-17-proof1}
	\end{align}
	Similarly, from (\ref{eq-15-proof1}), we will get:
	\begin{align}
		& \nonumber \emathbb{x_i|h^-_i}\exptsimp{\gobs_i|x_i, h^-_i}{\topo_Z}\int_{x_{i+1}}\prob{x_{i+1}|x_i, \policy^{\topo_Z}_i} r(x_{i+1}) dx_{i+1} \\ &\geq  \emathbb{x_i|h^-_i}\exptsimp{\gobs_i|x_i, h^-_i}{\topo_Z}\min_{a_i(x_i) \in \CA}\int_{x_{i+1}}\prob{x_{i+1}|x_i, a_i} r(x_{i+1}) dx_{i+1} \\
		& = \emathbb{x_i|h^-_i}\exptsimp{\gobs_i|x_i, h^-_i}{\topo}\min_{a_i(x_i) \in \CA}\int_{x_{i+1}}\prob{x_{i+1}|x_i, a_i} r(x_{i+1}) dx_{i+1}
		\label{eq-19-proof1}
	\end{align}
	Since we consider $\gobs^{\topo}_i=o_i$ at the belief node for the topology and the alternative observation model is fully observable, the actions in Equation (\ref{eq-17-proof1}) and (\ref{eq-19-proof1}) can be viewed as a policy: $a_i(x_i)=\policy_i, \policy_i \in \{\policy_i:\mathcal{H}^-_i \times \gobSpace_i^{\topo} \mapsto \aSpace\} := \Bar{\Pi}_i(\mathcal{H}^-_i, \gobSpace_i^{\topo}) $. 
	\begin{flalign}
		&\nonumber \emathbb{x_i|h^-_i}\exptsimp{\gobs_i|x_i, h^-_i}{\topo_Z}\int_{x_{i+1}}\prob{x_{i+1}|x_i, \policy^{\topo_Z}_i} r(x_{i+1}) dx_{i+1} \\ & \leq \max_{\bpolicy_i\in \Bar{\Pi}_i(\mathcal{H}^-_i, \gobSpace_i^{\topo})} \emathbb{x_i|h^-_i}\exptsimp{\gobs_i|x_i, h^-_i}{\topo}\int_{x_{i+1}}\prob{x_{i+1}|x_i, \bpolicy_i(h^-_i, \gobs^{\topo}_i)} r(x_{i+1}) dx_{i+1} 
	\end{flalign}
	\begin{flalign}
		& \nonumber \emathbb{x_i|h^-_i}\exptsimp{\gobs_i|x_i, h^-_i}{\topo_Z}\int_{x_{i+1}}\prob{x_{i+1}|x_i, \policy^{\topo_Z}_i} r(x_{i+1}) dx_{i+1} \\ & \geq \min_{\bpolicy_i\in \Bar{\Pi}_i(\mathcal{H}^-_i, \gobSpace_i^{\topo})} \emathbb{x_i|h^-_i}\exptsimp{\gobs_i|x_i, h^-_i}{\topo}\int_{x_{i+1}}\prob{x_{i+1}|x_i, \bpolicy_i(h^-_i, \gobs^{\topo}_i)} r(x_{i+1}) dx_{i+1} 
	\end{flalign}
	Since different policy $\bpolicy_i$ upper and lower bound the expected closed-loop reward, 
	we will also bound the distance between the expected closed-loop reward and a known value by exploring the policy space:
	\begin{align}
		\big|\emathbb{x_i|h^-_i}\exptsimp{\gobs_i|x_i, h^-_i}{\topo_Z} \int_{x_{i+1}}&\prob{x_{i+1}|x_i, \policy^{\topo_Z}_i} r(x_{i+1}) dx_{i+1} - \emathbb{x_i|h^{\topo-}_i}\exptsimp{\gobs_i|x_i, h^{\topo-}_i}{\topo}\emathbb{x_{i+1}|x_i, \policy^{\topo}_i}r(x_{i+1}) \big|  	 	\label{eq-22-proof1} \nonumber
		\\	\leq \max_{\bpolicy_i\in \Bar{\Pi}_i(\mathcal{H}^-_i, \gobSpace_i^{\topo})} &\big| \emathbb{x_i|h^-_i}\exptsimp{\gobs_i|x_i, h^-_i}{\topo}\int_{x_{i+1}}\prob{x_{i+1}|x_i, \bpolicy_i(h^-_i, \gobs^{\topo}_i)} r(x_{i+1}) dx_{i+1} \nonumber \\ &- \emathbb{x_i|h^{\topo-}_i}\exptsimp{\gobs_i|x_i, h^{\topo-}_i}{\topo}\emathbb{x_{i+1}|x_i, \policy^{\topo}_i}r(x_{i+1})  \big|
	\end{align}
	\textbf{Then, we consider the belief node uses the original observation space and model: $\beta(h^-_i)=1$}. 
	
	We have:
	\begin{equation}
		\emathbb{x_i|h^-_i}\exptsimp{\gobs_i|x_i, h^-_i}{\topo_Z}\emathbb{x_{i+1}|x_i, \policy^{\topo{Z}}_i}r(x_{i+1}) = \emathbb{x_i|h^-_i}\exptsimp{\gobs_i|x_i, h^-_i}{\topo}\emathbb{x_{i+1}|x_i, \policy^{\topo}_i}r(x_{i+1})
	\end{equation}
	Here, $\policy^{\topo{Z}}_i$ and $\policy^{\topo}_i$ share the same mapping, $\mathcal{H}^-_i\times\CZ_i\mapsto\CA$. We can say that the optimal and worst $\policy_i^{\topo}$ can upper and lower bound any $\policy^{\topo{Z}}_i$. Thus, there exists a policy $\bpolicy_i^{\topo}$ that can bound the distance in Equation (\ref{eq-22-proof1}).\qed
\end{proof}

If we take a further step from the above claim and assume the propagated history $h^-_{i}$ has the original POMDP topology $\topo_Z$, we will have:

	 \begin{align}
		\max_{\bpolicy_i\in \Bar{\Pi}_i(\mathcal{H}^-_i, \gobSpace_i^{\topo})} \big|\emathbb{x_i|h^-_i}&\exptsimp{\gobs_i|x_i, h^-_i}{\topo}\int_{x_{i+1}}\prob{x_{i+1}|x_i, \bpolicy_i^{\topo}}r(x_{i+1})\nonumber\\ & - \emathbb{x_i|h^{\topo-}_i} \exptsimp{\gobs_i|x_i, h^{\topo-}_i}{\topo}\emathbb{x_{i+1}|x_i, \policy^{\topo}_i}r(x_{i+1})\big| \\
		\triangleq \big|\emathbb{x_i|h^-_i}\exptsimp{\gobs_i|x_i, h^-_i}{\topo}\int_{x_{i+1}}&\prob{x_{i+1}|x_i, \bpolicystar_i(h^-_i,\gobs^{\topo}_i)}r(x_i) \nonumber\\ & - \emathbb{x_i|h^{\topo-}_i} \exptsimp{\gobs_i|x_i, h^{\topo-}_i}{\topo}\emathbb{x_{i+1}|x_i, \policy^{\topo}_i}r(x_{i+1})\big| \\
		= \big|\emathbb{x_{i-1}|h^-_{i-1}}\exptsimp{\gobs_{i-1}|x_{i-1}, h^-_{i-1}}{\topo_Z}&\emathbb{x_i|x_{i-1}, \policyc_{i-1}}\exptsimp{\gobs_i|x_i, h^-_i}{\topo}\int_{x_{i+1}}\prob{x_{i+1}|x_i, \bpolicystar_i(h^-_i,\gobs^{\topo}_i)}r(x_i)\nonumber\\& - \emathbb{x_i|h^{\topo-}_i} \exptsimp{\gobs_i|x_i, h^{\topo-}_i}{\topo}\emathbb{x_{i+1}|x_i, \policy^{\topo}_i}r(x_i)\big|
		\label{eq-26-proof2}
	\end{align}
	\textbf{i) If both $\beta(h_{i-1}^-)=0$ and $\beta(h_{i}^-)=0$}, 
	for every possible $x_i\in \CX$ and $x_{i-1} \in \CX$, we can find an action $a^{+}_{i}(x_{i})$ and $a^{+}_{i-1}(x_{i-1})$ such that:
	\begin{align}
		&\emathbb{x_i|x_{i-1}, \policyc_{i-1}}\int_{x_{i+1}}\prob{x_{i+1}|x_i, \bpolicystar_i}r(x_{i+1}) \nonumber \\ & \leq \max_{a_{i-1}(x_{i-1})} \emathbb{x_i|x_{i-1}, a_{i-1}(x_{i-1})}\max_{a_i(x_i)}\int_{x_{i+1}}\prob{x_{i+1}|x_i,a_i(x_i)}r(x_{i+1})
	\end{align}
	Then we have:
	\begin{align}
		\emathbb{x_{i-1}|h^-_{i-1}}\exptsimp{\gobs_{i-1}|x_{i-1}, h^-_{i-1}}{\topo_Z}&\emathbb{x_i|x_{i-1}, \policyc_{i-1}}\exptsimp{\gobs_i|x_i, h^-_i}{\topo}\int_{x_{i+1}}\prob{x_{i+1}|x_i, \bpolicystar_i(h^-_i,\gobs^{\topo}_i)}r(x_{i+1})\nonumber \\
		\leq \emathbb{x_{i-1}|h^-_{i-1}}\exptsimp{\gobs_{i-1}|x_{i-1}, h^-_{i-1}}{\topo_Z}& \max_{a_{i-1}(x_{i-1})} \emathbb{x_i|x_{i-1}, a_{i-1}(x_{i-1})}\exptsimp{\gobs_i|x_i, \{h_{i-1}, a_{i-1}\}}{\topo} \nonumber \\ &\max_{a_i(x_i)}\int_{x_{i+1}}\prob{x_{i+1}|x_i,a_i(x_i)}r(x_{i+1}) \\
		= \emathbb{x_{i-1}|h^-_{i-1}}\exptsimp{\gobs_{i-1}|x_{i-1}, h^-_{i-1}}{\topo} &\max_{a_{i-1}(x_{i-1})} \emathbb{x_i|x_{i-1}, a_{i-1}(x_{i-1})}\exptsimp{\gobs_i|x_i, \{h^-_{i-1}, z^{\topo}_{i-1}, a_{i-1}\}}{\topo} \nonumber\\ & \max_{a_i(x_i)}\int_{x_{i+1}}\prob{x_{i+1}|x_i,a_i(x_i)}r(x_{i+1})
	\end{align}

	Similarly, we can find the lower bound:
	\begin{align}
		\emathbb{x_{i-1}|h^-_{i-1}}\exptsimp{\gobs_{i-1}|x_{i-1}, h^-_{i-1}}{\topo_Z}\emathbb{x_i|x_{i-1}, \policyc_{i-1}}&\exptsimp{\gobs_i|x_i, h^-_i}{\topo}\int_{x_{i+1}}\prob{x_{i+1}|x_i, \bpolicystar_i(h^-_i,\gobs^{\topo}_i)}r(x_{i+1}) \\
		\geq \emathbb{x_{i-1}|h^-_{i-1}}\exptsimp{\gobs_{i-1}|x_{i-1}, h^-_{i-1}}{\topo_Z} &\min_{a_{i-1}(x_{i-1})} \emathbb{x_i|x_{i-1}, a_{i-1}(x_{i-1})}\exptsimp{\gobs_i|x_i,\{h_i, a_{i-1}\}}{\topo} \nonumber \\ & \min_{a_i(x_i)}\int_{x_{i+1}}\prob{x_{i+1}|x_i,a_i(x_i)}r(x_{i+1}) \\
		= \emathbb{x_{i-1}|h^-_{i-1}}\exptsimp{\gobs_{i-1}|x_{i-1}, h^-_{i-1}}{\topo} &\min_{a_{i-1}(x_{i-1})} \emathbb{x_i|x_{i-1}, a_{i-1}(x_{i-1})}\exptsimp{\gobs_i|x_i,\{h^-_i, z^{\topo}_{i-1}, a_{i-1}\}}{\topo} \nonumber \\ & \min_{a_i(x_i)}\int_{x_{i+1}}\prob{x_{i+1}|x_i,a_i(x_i)}r(x_{i+1})
	\end{align}
	The $a_{i-1}(x_{i-1})$ and $a_i(x_i)$ in the above equations can be viewed as policy respectively, $\policy_{i-1}\in \Pi_{i-1}$$(\mathcal{H}^-_{i-1},\gobSpace^{\topo}_{i-1})$ and $\policy_{i}\in\Pi_i(\mathcal{H}^-_{i-1},\gobSpace^{\topo}_{i-1}, \policy_{i-1}, \gobSpace^{\topo}_i)$.
	Similar to the previous claim, we can say there exists a policy to bound the below distance:
	\begin{align}
		\big|&\emathbb{x_{i-1}|h^-_{i-1}}\exptsimp{\gobs_{i-1}|x_{i-1}, h^-_{i-1}}{\topo_Z}\emathbb{x_i|x_{i-1}, \policyc_{i-1}}\exptsimp{\gobs_i|x_i, h^{-}_{i} }{\topo}\int_{x_{i+1}}\prob{x_{i+1}|x_i, \bpolicystar_i(h^{-}_i,\gobs^{\topo}_i)}r(x_{i+1})\\ & - \emathbb{x_i|h^{\topo-}_i} \exptsimp{\gobs_i|x_i, h^{\topo-}_i}{\topo}\emathbb{x_{i+1}|x_i, \policy^{\topo}_i}r(x_{i+1})\big| \nonumber\\ & \leq
		\max_{\bpolicy_{i-1}\in  \Bar{\Pi}_{i-1}(\mathcal{H}^-_{i-1},\gobSpace^{\topo}_{i-1}), \bpolicy_{i}(h^-_{i-1},z^{\topo}_{i-1}, \bpolicy_{i-1}(h^-_{i-1},z^{\topo}_{i-1}), z^{\topo}_i)}\nonumber\\& \big|\emathbb{x_{i-1}|h^-_{i-1}}\exptsimp{\gobs_{i-1}|x_{i-1}, h^-_{i-1}}{\topo}\emathbb{x_i|x_{i-1}, \bpolicy_{i-1}}\exptsimp{\gobs_i|x_i,h^{p-}_{i}}{\topo}\int_{x_{i+1}}\prob{x_{i+1}|x_i, \bpolicy_i}r(x_{i+1})\\& - \emathbb{x_i|h^{\topo-}_i} \exptsimp{\gobs_i|x_i, h^{\topo-}_i}{\topo}\emathbb{x_{i+1}|x_i, \policy^{\topo}_i}r(x_{i+1})\big|
		\label{eq-35-proof2}
	\end{align}

	ii) If $\beta(h_{i-1}^-)=0$ and $\beta(h_{i}^-)=1$, for each possible $x_{i-1} \in \CX$ and each $z_i \in \CZ_i$:
	\begin{align}
		&\emathbb{x_i|x_{i-1}, \policyc_{i-1}}\exptsimp{\gobs_i|x_i}{}\int_{x_{i+1}}\prob{x_{i+1}|x_i, \bpolicystar_i(h_{i-1}^-, z^{\topo_Z}_{i-1}, \policyc_{i-1},  z^{\topo}_i)}r(x_{i+1}) \nonumber \\ & \leq \max_{a_{i-1}(x_{i-1})} \emathbb{x_i|x_{i-1}, a_{i-1}(x_{i-1})}\exptsimp{\gobs_i|x_i}{}\max_{a_i(z_i)}\int_{x_{i+1}}\prob{x_{i+1}|x_i,a_i(z_i)}r(x_{i+1})
	\end{align}
	\begin{align}
		&\emathbb{x_i|x_{i-1}, \policyc_{i-1}}\exptsimp{\gobs_i|x_i}{}\int_{x_{i+1}}\prob{x_{i+1}|x_i, \bpolicystar_i(h_{i-1}^-, z^{\topo_Z}_{i-1}, \policyc_{i-1},  z^{\topo}_i)}r(x_{i+1}) \nonumber \\ & \geq \min_{a_{i-1}(x_{i-1})} \emathbb{x_i|x_{i-1}, a_{i-1}(x_{i-1})}\exptsimp{\gobs_i|x_i}{}\min_{a_i(z_i)}\int_{x_{i+1}}\prob{x_{i+1}|x_i,a_i(z_i)}r(x_{i+1})
	\end{align}
	The $a_{i-1}(x_{i-1})$ and $a_i(z_i)$ in the above equations can be viewed as policy respectively, $\policy_{i-1}(h^-_{i-1},z^{\topo}_{i-1})$ and $\policy_{i}(h^-_{i-1},z^{\topo}_{i-1}, \policy_{i-1}(h^-_{i-1},z^{\topo}_{i-1}), z^{\topo}_i)$. Then we can get the same result as Equation (\ref{eq-35-proof2}).
	
	iii) If both $\beta(h_{i-1}^-)=1$ and $\beta(h_{i}^-)=0$, the result is similar.
	
	iv) If $\beta(h_{i-1}^-)=1$ and $\beta(h_{i}^-)=1$, the policy $\bpolicy_{i-1,i}$ in Equation (\ref{eq-35-proof2}) shares the same mapping as the closed-loop policy $\policyc_{i-1,i}$. So there exists the optimal and worst policy that upper and lower bound the expected reward, so we can also get the same result as Equation (\ref{eq-35-proof2}).

\subsection{Main Proof of Lemma 1}
\begin{align}
	&\Big|\exptsimp{\gobs_{1:i}|b_0, \pi^{\topo}}{\topo}(r(b^{\topo}_{i})) -  \exptsimp{\gobs_{1:i}|b_0, \pi^{\topo_Z}}{\topo_Z}(r(b^{\topo_Z}_{i})) \Big|
	\intertext{Assume state-dependent reward}
	=& \Big|\exptsimp{\gobs_{1:i}|b_0, \pi^{\topo}}{\topo} \emathbb{x_i|b^{\topo}_i}r(x_i) -  \exptsimp{\gobs_{1:i}|b_0, \pi^{\topo_Z}}{\topo_Z}\emathbb{x_i|b^{\topo_Z}_i}(r(x_{i})) \Big|\\
	= & \Big|\exptsimp{\gobs_{1:i-1}|b_0, \pi^{\topo}}{\topo} \emathbb{x_i|h^{\topo-}_i}\exptsimp{\gobs_i|x_i, h^{\topo-}_i}{\topo}r(x_i) -  \exptsimp{\gobs_{1:i-1}|b_0, \pi^{\topo_Z}}{\topo_Z}\emathbb{x_i|h^{\topo_Z- }_i}\exptsimp{\gobs_i|x_i}{\topo_Z}(r(x_{i})) \Big|
	\intertext{(Chain rule and cancel $z_i$)}
	=& \Big|\exptsimp{\gobs_{1:i-1}|b_0, \pi^{\topo}}{\topo} \emathbb{x_i|h^{\topo-}_i}r(x_i) -  \exptsimp{\gobs_{1:i-1}|b_0, \pi^{\topo_Z}}{\topo_Z}\emathbb{x_{i-1}|h^{\topo_Z}_{i-1}}\emathbb{x_i|x_{i-1}, \policyc_{i-1}}(r(x_i)) \Big| \\  \intertext{(Bayes Rule)} 
	= &  \Big|\exptsimp{\gobs_{1:i-1}|b_0, \pi^{\topo}}{\topo} \emathbb{x_i|h^{\topo-}_i}r(x_i) -  \exptsimp{\gobs_{1:i-2}|b_0, \policy^{\topo_Z}}{\topo_Z}\emathbb{x_{i-1}|h^{\topo_Z-}_{i-1}}\exptsimp{\gobs_{i-1}|x_{i-1}, h^{\topo_Z-}_{i-1}}{\topo_Z} \emathbb{x_i|x_{i-1},\policy^{\topo_Z}_{i-1}}r(x_i) \Big|
	\label{equ:state-dependent-and-chain-rule}\\
	=	&\Big|\exptsimp{\gobs_{1:i-1}|b_0, \pi^{\topo}}{\topo} \emathbb{x_i|h^{\topo-}_i}r(x_i) \nonumber \\ &-\emathbb{x_0|b_0}\emathbb{x_1|x_0,\policyc_0}\exptsimp{\gobs_1|x_1, h^{p-}_{1}}{\topo_Z}...\emathbb{x_{i-1}|x_{i-2},\policyc_{i-2}} \exptsimp{\gobs_{i-1}|x_{i-1},h^{p-}_{i-1}}{\topo_Z}\emathbb{x_i|x_{i-1},\policy^{\topo_Z}_{i-1}}r(x_i) \Big| \\
	\intertext{Here,  $\bar{h}^{\topo-}$ is propagated history with topology $\topo$. It can be easily got by just repeating the process in Section \ref{pre-proof} from time $t=i$ to the initial time $t=0$.}
	 \leq& \max_{\bpolicy^{\topo}}\Big|\exptsimp{\gobs_{1:i-1}|b_0, \pi^{\topo}}{\topo} \emathbb{x_i|h^{\topo-}_i}r(x_i) \nonumber\\&-\emathbb{x_0|b_0}\emathbb{x_1|x_0,\bpolicy^{\topo}_0}\exptsimp{\gobs_1|x_1,\bar{h}^{\topo-}_{1}}{\topo}...\emathbb{x_{i-1}|x_{i-2},\bpolicy^{\topo}_{i-2}} \exptsimp{\gobs_{i-1}|x_{i-1},\bar{h}^{\topo-}_{i}}{\topo}\emathbb{x_i|x_{i-1},\bpolicy^{\topo}_{i-1}}r(x_i) \Big|
\end{align} 

\section{Proof of Lemma \ref{lemma:bound-q}}\label{sec:Proof_lemma:bound-q}
\begin{proof}
	We can look into the difference in the Q function between two different topologies, which is defined as below:
	\begin{align}
		& \Delta Q(b_0,\action_0, \policy^{\topo_Z},\policy^{\topo},\topo_Z, \topo) \triangleq| Q_{\topo_Z}^{{\policy^{\topo_Z}}}(b_0,\action_0) - Q_{\topo}^{\policy^{\topo}}(b_0,\action_0) | \\ 
		= & \big| \sum_{i=1}^{L} \exptsimp{\gobs_{1:i}|b_0,\policy^{\topo_Z}}{\topo_Z}(r({\tb}_{i})) - \sum_{i=1}^{L} \exptsimp{\gobs_{1:i}|b_0,\policy^{\topo}}{\topo}(r(b_{i})) \big| \\
		\intertext{(Assume state-dependent reward and chain rule, same as Equation (\ref{equ:state-dependent-and-chain-rule}))}
		= & \Big|\sum_{i=1}^{L}\big[\exptsimp{\gobs_{1:i-2}|b_0,\policy^{\topo_Z}}{\topo_Z}\emathbb{x_{i-1}|h^{\topo_Z-}_{i-1}}\exptsimp{\gobs_{i-1}|x_{i-1}, h^{\topo_Z-}_{i-1}}{\topo_Z} \emathbb{x_i|x_{i-1},\policy^{\topo_Z}_{i-1}}r(x_i)\ \nonumber \\& -  \exptsimp{\gobs_{1:i-2}|b_0,\policy^{\topo}}{\topo}\emathbb{x_{i-1}|h^{\topo-}_{i-1}}\exptsimp{\gobs_{i-1}|x_{i-1}, h^{\topo-}_{i-1}}{\topo} \emathbb{x_i|x_{i-1},\policy^{\topo}_{i-1}}r(x_i)\ \big]\Big| \\
		\intertext{(Take out the expected reward $r(x_L)$ from the summation)}
		= & \nonumber \Big|\sum_{i=1}^{L-1}\big[\exptsimp{\gobs_{1:i-2}|b_0,\policy^{\topo_Z}}{\topo_Z}\emathbb{x_{i-1}|h^{\topo_Z-}_{i-1}}\exptsimp{\gobs_{i-1}|x_{i-1}, h^{\topo_Z-}_{i-1}}{\topo_Z} \emathbb{x_i|x_{i-1},\policy^{\topo_Z}_{i-1}}r(x_i)\ \nonumber \\ &-  \exptsimp{\gobs_{1:i-2}|b_0,\policy^{\topo}}{\topo}\emathbb{x_{i-1}|h^{\topo-}_{i-1}}\exptsimp{\gobs_{i-1}|x_{i-1}, h^{\topo-}_{i-1}}{\topo} \emathbb{x_i|x_{i-1},\policy^{\topo}_{i-1}}r(x_i)\ \big] \nonumber \\ &  +  \exptsimp{\gobs_{1:L-2}|b_0,\policy^{\topo_Z}}{\topo_Z}\emathbb{x_{L-1}|h^{\topo_Z-}_{L-1}}\exptsimp{\gobs_{L-1}|x_{L-1}, h^{p-}_{L-1}}{\topo_Z} \emathbb{x_L|x_{L-1},\policy^{\topo_Z}_{L-1}}r(x_L)\ \nonumber \\ &-  \exptsimp{\gobs_{1:L-2}|b_0,\policy^{\topo}}{\topo}\emathbb{x_{L-1}|h^{\topo-}_{L-1}}\exptsimp{\gobs_{L-1}|x_{L-1}, h^{\topo-}_{L-1}}{\topo} \emathbb{x_L|x_{L-1},\policy^{\topo}_{L-1}}r(x_L)\ \Big| \\
		\intertext{(Bound the expected reward $r(x_L)$ using the same method in Section \ref{pre-proof})}  
		\leq & \nonumber \max_{\bpolicy^{\topo}_{L-1}\in\Bar{\Pi}_{L-1}(\mathcal{H}^{\topo_Z-}_{L-1}, \gobSpace^{\topo}_{L-1})} \Big|\sum_{i=1}^{L-1}\big[\exptsimp{\gobs_{1:i-2}|b_0,\policy^{\topo_Z}}{\topo_Z}\emathbb{x_{i-1}|h^{\topo_Z-}_{i-1}}\exptsimp{\gobs_{i-1}|x_{i-1}, h^{\topo_Z-}_{i-1}}{\topo_Z} \emathbb{x_i|x_{i-1},\policy^{\topo_Z}_{i-1}}r(x_i)\ \nonumber \\ &-  \exptsimp{\gobs_{1:i-2}|b_0,\policy^{\topo}}{\topo}\emathbb{x_{i-1}|h^{\topo-}_{i-1}}\exptsimp{\gobs_{i-1}|x_{i-1}, h^{\topo-}_{i-1}}{\topo} \emathbb{x_i|x_{i-1},\policy^{\topo}_{i-1}}r(x_i)\ \big] \nonumber \\ &  +
		\exptsimp{\gobs_{1:L-2}|b_0,\policy^{\topo_Z}}{\topo_Z}\emathbb{x_{L-1}|h^{\topo_Z-}_{L-1}}\exptsimp{\gobs_{L-1}|x_{L-1}, h^{\topo_Z-}_{L-1}}{\topo} \emathbb{x_L|x_{L-1},\bpolicy^{\topo}_{L-1}}r(x_L)\ \nonumber \\ &-  \exptsimp{\gobs_{1:L-2}|b_0,\policy^{\topo}}{\topo}\emathbb{x_{L-1}|h^{\topo-}_{L-1}}\exptsimp{\gobs_{L-1}|x_{L-1}, h^{\topo-}_{L-1}}{\topo} \emathbb{x_L|x_{L-1},\policy^{\topo}_{L-1}}r(x_L)\ \Big | \\
		\intertext{(Take out the expected reward $r(x_{L-1})$ from the summation, and denote the policy maximizing the distance as $\bpolicy^{+}_{L-1}$)} 
		=& \nonumber  \Big|\sum_{i=1}^{L-2}\big[\exptsimp{\gobs_{1:i-2}|b_0,\policy^{\topo_Z}}{\topo_Z}\emathbb{x_{i-1}|h^{\topo_Z-}_{i-1}}\exptsimp{\gobs_{i-1}|x_{i-1}, h^{\topo_Z-}_{i-1}}{\topo_Z} \emathbb{x_i|x_{i-1},\policy^{\topo_Z}_{i-1}}r(x_i)\ \nonumber \\ &-  \exptsimp{\gobs_{1:i-2}|b_0,\policy^{\topo}}{\topo}\emathbb{x_{i-1}|h^{\topo-}_{i-1}}\exptsimp{\gobs_{i-1}|x_{i-1}, h^{\topo-}_{i-1}}{\topo} \emathbb{x_i|x_{i-1},\policy^{\topo}_{i-1}}r(x_i)\ \big] 
		\\ \nonumber & \quad + \exptsimp{\gobs_{1:L-3}|b_0,\policy^{\topo_Z}}{\topo_Z}\emathbb{x_{L-2}|h^{\topo_Z-}_{L-2}}\exptsimp{\gobs_{L-2}|x_{L-2}, h^{\topo_Z-}_{L-2}}{\topo_Z} \emathbb{x_{L-1}|x_{L-2},\policy^{\topo_Z}_{L-2}}r(x_{L-1})\ \nonumber \\ &-  \exptsimp{\gobs_{1:L-3}|b_0,\policy^{\topo}}{\topo}\emathbb{x_{L-2}|h^{\topo-}_{L-2}}\emathbb{\gobs_{L-2}|h^{\topo-}_{L-2},\topo} \emathbb{x_{L-1}|x_{L-2},\policy^{\topo}_{L-2}}r(x_{L-1})
		\\ & \quad  +
		\exptsimp{\gobs_{1:L-2}|b_0,\policy^{\topo_Z}}{\topo_Z}\emathbb{x_{L-1}|h^{\topo_Z-}_{L-1}}\exptsimp{\gobs_{L-1}|x_{L-1}, h^{\topo_Z-}_{L-1}}{\topo} \emathbb{x_L|x_{L-1},\bpolicy^{+}_{L-1}}r(x_L)\ \nonumber \\ &-  \exptsimp{\gobs_{1:L-2}|b_0,\policy^{\topo}}{\topo}\emathbb{x_{L-1}|h^{\topo-}_{L-1}}\exptsimp{\gobs_{L-1}|x_{L-1}, h^{\topo-}_{L-1}}{\topo} \emathbb{x_L|x_{L-1},\policy^{\topo}_{L-1}}r(x_L)\ \Big | 
		\\ = 
		\nonumber &  \Big|\sum_{i=1}^{L-2}\big[\exptsimp{\gobs_{1:i-2}|b_0,\policy^{\topo_Z}}{\topo_Z}\emathbb{x_{i-1}|h^{\topo_Z-}_{i-1}}\exptsimp{\gobs_{i-1}|x_{i-1}, h^{\topo_Z-}_{i-1}}{\topo_Z} \emathbb{x_i|x_{i-1},\policy^{\topo_Z}_{i-1}}r(x_i) \nonumber \\ & -  \exptsimp{\gobs_{1:i-2}|b_0,\policy^{\topo}}{\topo}\emathbb{x_{i-1}|h^{\topo-}_{i-1}}\exptsimp{\gobs_{i-1}|x_{i-1}, h^{\topo-}_{i-1}}{\topo} \emathbb{x_i|x_{i-1},\policy^{\topo}_{i-1}}r(x_i) \big] 
		\\ \nonumber &  + \exptsimp{\gobs_{1:L-3}|b_0,\policy^{\topo_Z}}{\topo_Z}\emathbb{x_{L-2}|h^{\topo_Z-}_{L-2}}\exptsimp{\gobs_{L-2}|x_{L-2}, h^{\topo_Z-}_{L-2}}{\topo_Z} \emathbb{x_{L-1}|x_{L-2},\policy^{\topo_Z}_{L-2}}\Big(r(x_{L-1}) \nonumber \\ & \qquad +\exptsimp{\gobs_{L-1}|x_{L-1}, h^{\topo_Z-}_{L-1}}{\topo} \emathbb{x_L|x_{L-1},\bpolicy^{+}_{L-1}(h^{\topo_Z-}_{L-1}, \gobs^{\topo}_{L-1})}r(x_{L})\Big) 
		\\ &  -
		\exptsimp{\gobs_{1:L-3}|b_0,\policy^{\topo}}{\topo}\emathbb{x_{L-2}|h^{\topo-}_{L-2}}\exptsimp{\gobs_{L-2}|x_{L-2}, h^{\topo-}_{L-2}}{\topo} \emathbb{x_{L-1}|x_{L-2},\policy^{\topo}_{L-2}}r(x_{L-1})\nonumber \\ &	-  \exptsimp{\gobs_{1:L-2}|b_0,\policy^{\topo}}{\topo}\emathbb{x_{L-1}|h^{\topo-}_{L-1}}\exptsimp{\gobs_{L-1}|x_{L-1}, h^{\topo-}_{L-1}}{\topo} \emathbb{x_L|x_{L-1},\policy^{\topo}_{L-1}}r(x_L)\ \Big | 			
		\intertext{(Bound the term $\emathbb{x_{L-1}|x_{L-2},\policy^{\topo_Z}_{L-2}}(r(x_{L-1}) +\exptsimp{\gobs_{L-1}|x_{L-1}, h^{\topo_Z-}_{L-1}}{\topo} \emathbb{x_L|x_{L-1},\bpolicy^{+}_{L-1}}r(x_{L}))$ using a similar method as Section \ref{pre-proof})} 
		\leq
		&\nonumber\max_{\bpolicy_{L-2,L-1}} \Big|\sum_{i=1}^{L-2}\big[\exptsimp{\gobs_{1:i-2}|b_0,\policy^{\topo_Z}}{\topo_Z}\emathbb{x_{i-1}|h^{\topo_Z-}_{i-1}}\exptsimp{\gobs_{i-1}|x_{i-1},h^{\topo_Z-}_{i-1}}{\topo_Z} \emathbb{x_i|x_{i-1},\policy^{\topo_Z}_{i-1}}r(x_i) \nonumber \\ & \qquad \ -  \exptsimp{\gobs_{1:i-2}|b_0,\policy^{\topo}}{\topo}\emathbb{x_{i-1}|h^{\topo-}_{i-1}}\exptsimp{\gobs_{i-1}|x_{i-1}, h^{\topo-}_{i-1}}{\topo} \emathbb{x_i|x_{i- 1},\policy^{\topo}_{i-1}}r(x_i) \big] 
		\\ &\nonumber \qquad \ + \emathbb{\gobs_{1:L-3}|,b_0,\policy^{\topo_Z},\topo_Z}\emathbb{x_{L-2}|h^{\topo_Z-}_{L-2}}\exptsimp{\gobs_{L-2}|x_{L-2}, h^{\topo_Z-}_{L-2}}{\topo} \emathbb{x_{L-1}|x_{L-2},\bpolicy_{L-2}}\Big(r(x_{L-1}) \nonumber \\ & \qquad \qquad +\exptsimp{\gobs_{L-1}|x_{L-1}, h^{p-}_{L-1}}{\topo} \emathbb{x_L|x_{L-1},\bpolicy_{L-1}}r(x_{L})\Big) 
		\\ & \qquad  \ -
		\exptsimp{\gobs_{1:L-3}|b_0,\policy^{\topo}}{\topo}\emathbb{x_{L-2}|h^{\topo-}_{L-2}}\exptsimp{\gobs_{L-2}|x_{L-2}, h^{\topo-}_{L-2}}{\topo} \emathbb{x_{L-1}|x_{L-2},\policy^{\topo}_{L-2}}r(x_{L-1}) \nonumber \\ & \qquad \	-  \exptsimp{\gobs_{1:L-2}|b_0,\policy^{\topo}}{\topo}\emathbb{x_{L-1}|h^{\topo-}_{L-1}}\exptsimp{\gobs_{L-1}|x_{L-1}, h^{\topo-}_{L-1}}{\topo} \emathbb{x_L|x_{L-1},\policy^{\topo}_{L-1}}r(x_L)\ \Big | 			\\
		\intertext{(We can do it in a recursive way)} & \leq \max_{\bpolicy^{\topo}\in \Pi^{\topo}} | Q_{\topo_Z}^{{\bpolicy^{\topo}}}(b_0,\action_0) - Q_{\topo}^{\policy^{\topo}}(b_0,\action_0) |
	\end{align}     
	
	Which completes the proof of Lemma \ref{lemma:bound-q}. 
\end{proof}

\section{Proof of Theorem \ref{theorem:bound-q}}
The proof can be simple.

\emph{Proof:}
\begin{align}		Q_{\topo'}^{\policy^{\topo_Z}}(b_k,\action_k)  & \leq \min_{\policy^{\topo}\in \Pi^{\topo}} [Q_{\topo}^{\policy^{\topo}}(b_k,\action_k) +\delta_Q(b_k,\action_k, \policy^{\topo},\topo) ]\\ & = \min_{\policy^{\topo}\in \Pi^{\topo}} [Q_{\topo}^{\policy^{\topo}}(b_k,\action_k)+\max_{\bpolicy^{\topo}\in \Pi^{\topo}} | Q_{\topo}^{{\bpolicy^{\topo}}}(b_k,\action_k) - Q_{\topo}^{\policy^{\topo}}(b_k,\action_k) |] \label{theorem-q-equ1}
	\\
	& = \max_{\policy^{\topo}\in \Pi^{\topo}}[Q_{\topo}^{\policy^{\topo}}(b_k,\action_k)]\label{theorem-q-equ2}
\end{align}
The second direction in \eqref{eq:BoundQ} can be proved similarly. 
\qed
\section{Proof of Theorem 2}
\begin{proof}
	Let's consider the end of the planning horizon $L$.
	
	If $\simpfunc^{\topo}(\shis(b^-_{L-1}))=1$, we have:
	\begin{flalign}
		\int_{z_{L-1}}p(z_{L-1}|x_{L-1})&\max_{\policy^{\topo}_{L-1}\in\Pi^{\topo}}\int_{x_L}p(x_L|x_{L-1},\policy^{\topo}_{L-1})r(x_L) \\  & = \int_{z_{L-1}}p(z_{L-1}|x_{L-1})\int_{x_L}p(x_L|x_{L-1},\policy^{\topo_Z*}_{L-1})r(x_L) 
		\label{equ:opti-1}
	\end{flalign}
	If $\simpfunc^{\topo}(\shis(b^-_{L-1}))=0$, we have:
	\begin{flalign}
		\min_{\policy^{\topo}_{L-1}\in\Pi^{\topo}}\int_{x_L}&p(x_L|x_{L-1},\policy^{\topo}_{L-1})r(x_L) \nonumber \\ \leq \int_{z_{L-1}}&p(z_{L-1}|x_{L-1})\int_{x_L}p(x_L|x_{L-1},\policy^{\topo_Z*}_{L-1}) r(x_L) 
		\label{equ:opti-2}
	\end{flalign}
	We denote the left-hand side in Equation (\ref{equ:opti-1}) and (\ref{equ:opti-2}) as $LHS1(x_{L-1}, \topo)$ and the right-hand side as $RHS1(x_{L-1}, \topo_Z)$. In general, we can say $LHS1(x_{L-1}, \topo) \leq RHS1(x_{L-1}, \topo_Z)$. 
	
	Then, we take one more step earlier. Also if
	$\simpfunc^{\topo}(\shis(b^-_{L-2}))=1$
	, we have:
	\begin{flalign}
		&\nonumber\int_{z_{L-1}}p(z_{L-1}|x_{L-1})
		\max_{\policy_{L-2}^{\topo}\in\Pi^{\topo}}\int_{x_{L-1}}p(x_{L-1}|x_{L-2},\policy_{L-2}^{\topo})[r(x_{L-1})+LHS1(x_{L-1}, \topo)] \\
		& \leq \int_{z_{L-2}}p(z_{L-2}|h_{L-2})\int_{x_{L-2}}p(x_{L-2}|x_{L-2},\policy_{L-2}^{\topo_Z*})[r(x_{L-1})+RHS1(x_{L-1}, \topo)]
		\label{equ:opti-3}
	\end{flalign}
	And if $\simpfunc^{\topo}(\shis(b^-_{L-2}))=0$, we have:
	\begin{flalign}
		&\min_{\policy_{L-2}^{\topo}\in\Pi^{\topo}}\int_{x_{L-1}}p(x_{L-1}|x_{L-2},\policy_{L-2}^{\topo})[r(x_{L-1})+LHS1(x_{L-1}, \topo)]\nonumber \\
		& \leq \int_{z_{L-2}}p(z_{L-2}|h_{L-2})\int_{x_{L-2}}p(x_{L-2}|x_{L-2},\policy_{L-2}^{\topo_Z*})[r(x_{L-1})+RHS1(x_{L-1}, \topo)]
		\label{equ:opti-4}
	\end{flalign}
	We also denote the left-hand side of the inequality in Equation (\ref{equ:opti-3}) and (\ref{equ:opti-4}) as $LHS2(x_{L-2}, \topo)$ and the right-hand side as $RHS2(x_{L-2}, \topo_Z)$. In general, we can say $LHS2(x_{L-2}, \topo) \leq RHS2(x_{L-2}, \topo_Z)$. 
	
	The left-hand side is the iterative way to calculate $lb(b_t,\policy_t,\topo)$. If we keep iterate the above process from the end to the beginning of the planning horizon, we will get the inequality at the top of the belief tree:  $ lb(b_0, \action_0, \topo) \leq Q_{\topo_Z}^{\policy^{\topo_Z*}}(b_0,\action_0) $.

\end{proof}

\section{Sparse Sampling}
For a random variable $X\sim \mathcal{P}$, we use another sampling-based random variable, $Y\sim \mathcal{Q}$ to approximate the expectation of the original theoretic value. If we obtain some samples $\{y_i\}_{i=1}^N$ from a known distribution $\mathcal{Q}$, the calculation will be: 
\begin{equation}
\emathbb{X\sim \mathcal{P}}[f(X)] = \int f(x)\mathcal{P}(x)dx = \int f(x)\frac{\mathcal{P}(x)}{\mathcal{Q}(x)}\mathcal{Q}(x)dx \simeq \frac{1}{N}\sum_{i=1}^{N}\frac{\mathcal{P}(y_i)}{\mathcal{Q}(y_i)} f(y_i)
\end{equation}

Usually, the proposed distribution $\mathcal{Q}$ is the known motion model. 
The weighted sampling can be more efficient, where resampling will create more particles from the samples with higher weights. The belief density is approximated by $b(x)\simeq \frac{\sum_{i=1}^N w^i \delta(x-x^i)}{\sum_{i=1}^N w^i}$.

\subsection{Proof of Theorem \ref{theorem:upper-bound-estimation}}
\label{proof: upper-bound-estimation}
\begin{theorem}
	[Theorem \ref{theorem:upper-bound-estimation}] \textbf{Bounded Estimation Error.} For all the depth $d=0,...,L-1$ and $a$, the following concentration bound holds with probability at least $1-2|A|(|A|C)^L\exp(\frac{-C\lambda^2}{2V_{\max}^2})$ :
	\begin{equation}
	\Delta \hat{ub}(b_0,a_0, \topo) \leq \frac{L(L-1)}{2}\lambda, \qquad
	\Delta \hat{lb}(b_0,a_0, \topo) \leq \frac{L(L-1)}{2}\lambda.
\end{equation}
\end{theorem}

For the upper bound:
\begin{proof}
    We will follow a similar way to prove it as \cite{Kearns02ml, Lim23jair}.

    Firstly, we consider a general situation at a depth of $d$ and try to bound the error of estimating the optimal Q function given a belief $b_d$ and an action $a_d$:
    \begin{flalign}
       \delta^{\topo}_{\Hat{ub}}(d) &\triangleq |{Q}_{\topo}^{\policy^{\topo*}}(b_d,\action_d)-\hat{Q}_{\topo}^{\policy^{\topo*}}(b_d,\action_d) | \\& = \Big|r(b_d,\action_d) +  \emathbb{b_{d+1}|b_d,\action_d}[V^{\topo*}(b_{d+1})] - r(b_d, \action_d) - \frac{1}{C}\sum_{i=1}^{C}\hat{V}^{\topo*}(b_{d+1}'^{[I_i]})\Big|\\
        &=\Big|\emathbb{z_{d+1}|b_d,\action_d,\topo}[V^{\topo*}(b_{d+1})] - \frac{1}{C}\sum_{i=1}^{C}\hat{V}^{\topo*}(\Bar{b}_{d+1}^{\topo, [I_i]})\Big| \\
        & \leq \Big| \emathbb{z_{d+1}|b_d,\action_d,\topo}[V^{\topo*}(b_{d+1})] -  \frac{1}{C}\sum_{i=1}^{C}V^{\topo*}(\Bar{b}_{d+1}^{\topo, [I_i]}) \Big|+\Big| \frac{1}{C}\sum_{i=1}^{C}V^{\topo*}(\Bar{b}_{d+1}^{\topo, [I_i]}) -  \frac{1}{C}\sum_{i=1}^{C}\hat{V}^{\topo*}(\Bar{b}_{d+1}^{\topo, [I_i]}) \Big| \label{equ:esti-1}
    \end{flalign}

Here, the propagated next step belief state samples for a given belief tree topology $\topo$ is denoted as $\Bar{b}_{d+1}^{\topo, [I_i]}$. It is updated by the motion and observation models for the given topology $\topo$:
\begin{equation}
     \Bar{b}_{d+1}^{\topo, [I_i]} \sim
     \psi^{\topo}(\Bar{b}_{d+1}^{\topo} | b_d, \action_d) =
     \indicator_{\simpfunc(h^-_{d+1})=1}\alpha\cdot \prob{z_{d+1}|x_{d+1}}\prob{x_{d+1}|b_d, a_d} + \indicator_{\simpfunc(h^-_{d+1})=0}\prob{x_{d+1}|b_d, a_d}
\end{equation}
Here, $\alpha$ is a Bayes normalization factor.
For the first term in Equation (\ref{equ:esti-1}), we can directly use the Hoeffding's inequality:
\begin{flalign}
\prob{\Big|\emathbb{z_{d+1}|b_d,\action_d,\topo}[V^{\topo*}(b_{d+1})] -  \frac{1}{C}\sum_{i=1}^{C}V^{\topo*}(b_{d+1}'^{[I_i]})\Big| \leq \lambda } \geq 1- 2 \exp(\frac{-C\lambda^2}{2V_{\max}^2})
\end{flalign}

For the second term in Equation (\ref{equ:esti-1}), we analyze it in an iterative way:
\begin{flalign}
    \Big| \frac{1}{C}\sum_{i=1}^{C}V^{\topo*}(\Bar{b}_{d+1}^{\topo, [I_i]}) -  \frac{1}{C}\sum_{i=1}^{C}\hat{V}^{\topo*}(\Bar{b}_{d+1}^{\topo, [I_i]}) \Big| &= \frac{1}{C}\Big| \sum_{i=1}^{C}Q_{\topo}^{\policy^{\topo*}}(\Bar{b}_{d+1}^{\topo, [I_i]}, \action^*_{d+1})- \sum_{i=1}^{C}\hat{Q}_{\topo}^{\policy^{\topo*}}(\Bar{b}_{d+1}^{\topo, [I_i]}, \action^*_{d+1}) \Big| \\
    &\leq \delta^{\topo}_{\Hat{ub}}(d+1)
\end{flalign}

So, the below iterative bound is satisfied with a probability of at least $1-2|A|(|A|C)^{L-d}\exp(\frac{-C\lambda^2}{2V_{\max}^2})$:
\begin{flalign}
    \delta^{\topo}_{\Hat{ub}}(d) \leq \lambda +  \delta^{\topo}_{\Hat{ub}}(d+1)
    \label{equ:esti-2}
\end{flalign}
Here, the probability is based on the worst case for iteration, where it requires all the child belief nodes generated are well estimated with the number of $|A|C$ child nodes for each time step.

The estimation error at the end of the planning horizon $L-1$ is: 
\begin{equation}
    \delta^{\topo}_{\Hat{ub}}(L-1) = \lambda
\end{equation}

Then, we can find out the estimation bound at the top of the given belief tree by calculating it from the bottom to the top by Equation (\ref{equ:esti-2}).
\end{proof}

For the estimation of lower bound:
\begin{proof}
	Firstly, we consider a general situation at a depth of $d$ and try to bound the error of estimating the lower bound function given a belief $b_d$ and an action $a_d$:
	\begin{flalign}
		\delta^{\topo}_{lb}(t) \triangleq|&{lb}(b_t,a_t, \topo) - \hat{lb}(b_t,a_t, \topo)| \\=\Big|&\indicator_{\simpfunc(\shis(b^-_{t+1}))=1}\big[\emathbb{z_{t+1}|\shis(b_{t}),\action_t,\topo}\max_{\policy_{t+1}}lb(b_{t+1},\policy_{t+1},\topo)- \frac{1}{C}\sum_{i=1}^{C}\max_{\policy_{t+1}}\hat{lb}(\Bar{b}^{I_i}_{t+1},\policy_{t+1},\topo)\big]\nonumber \\
		+ &\indicator_{\simpfunc(\shis(b^-_{t+1}))=0}\big[\emathbb{z_{t+1}|\shis(b_{t}),\action_t,\topo} \min_{\policy_{t+1}}lb(b_{t+1},\policy_{t+1},\topo)-\frac{1}{C}\sum_{i=1}^{C} \min_{\policy_{t+1}}\hat{lb}(\Bar{b}^{I_i}_{t+1},\policy_{t+1},\topo)\big]\Big|\\
		\leq \Big|&\indicator_{\simpfunc(\shis(b^-_{t+1}))=1}\{\big| \emathbb{z_{t+1}|\shis(b_{t}),\action_t,\topo}\max_{\policy_{t+1}}lb(b_{t+1},\policy_{t+1},\topo) - \frac{1}{C}\sum_{i=1}^{C}\max_{\policy_{t+1}}lb(\Bar{b}^{I_i}_{t+1},\policy_{t+1},\topo) \big|\nonumber
		\\ &+ \ \big|\frac{1}{C}\sum_{i=1}^{C}\max_{\policy_{t+1}}lb(\Bar{b}^{I_i}_{t+1},\policy_{t+1},\topo) - \frac{1}{C}\sum_{i=1}^{C}\max_{\policy_{t+1}}\hat{lb}(\Bar{b}^{I_i}_{t+1},\policy_{t+1},\topo)\big| \} \nonumber
		\\
		+ &\indicator_{\simpfunc(\shis(b^-_{t+1}))=0}   \{\big| \emathbb{z_{t+1}|\shis(b_{t}),\action_t,\topo}\min_{\policy_{t+1}}lb(b_{t+1},\policy_{t+1},\topo) - \frac{1}{C}\sum_{i=1}^{C}\min_{\policy_{t+1}}lb(\Bar{b}^{I_i}_{t+1},\policy_{t+1},\topo) \big| \nonumber
		\\ &+ \big|\frac{1}{C}\sum_{i=1}^{C}\min_{\policy_{t+1}}lb(\Bar{b}^{I_i}_{t+1},\policy_{t+1},\topo) - \frac{1}{C}\sum_{i=1}^{C}\min_{\policy_{t+1}}\hat{lb}(\Bar{b}^{I_i}_{t+1},\policy_{t+1},\topo) \big|\} \Big| \label{equ:lb-lambda-pre}
	\end{flalign}
	Here, we can use a technique similar to the previous proof. We can directly use the Hoeffding inequality to bound the first term in each indicator as:
	\begin{equation}
		\prob{\Big|\emathbb{z_{t+1}|\shis(b_{t}),\action_t,\topo}\max_{\policy_{t+1}}lb(b_{t+1},\policy_{t+1},\topo) - \frac{1}{C}\sum_{i=1}^{C}\max_{\policy_{t+1}}lb(\Bar{b}^{I_i}_{t+1},\policy_{t+1},\topo)\Big| \leq \lambda } \geq 1- 2 \exp(\frac{-C\lambda^2}{2V_{\max}^2})
	\end{equation}
	\begin{equation}
		\prob{\Big|
			\emathbb{z_{t+1}|\shis(b_{t}),\action_t,\topo}\min_{\policy_{t+1}}lb(b_{t+1},\policy_{t+1},\topo) - \frac{1}{C}\sum_{i=1}^{C}\min_{\policy_{t+1}}lb(\Bar{b}^{I_i}_{t+1},\policy_{t+1},\topo)
			\Big| \leq \lambda } \geq 1- 2 \exp(\frac{-C\lambda^2}{2V_{\max}^2})
	\end{equation}
 The second term is also iteratively bounded by  $\delta^{\topo}_{lb}(t+1)$.
	So, combining the two probabilistic bounds together, we can get the iterative bound for the estimator satisfied with a probability of at least $1-2|A|(|A|C)^{L-ts}\exp(\frac{-C\lambda^2}{2V_{\max}^2})$:
	\begin{flalign}
		\delta^{\topo}_{lb}(t)\leq (\ref{equ:lb-lambda-pre})\leq &\indicator_{\simpfunc(\shis(b^-_{t+1}))=1} \big[\lambda + \delta^{\topo}_{lb}(t+1)\big] + \indicator_{\simpfunc(\shis(b^-_{t+1}))=0} \big[\lambda + \delta^{\topo}_{lb}(t+1)\big]\label{equ:lb-lambda} \\
		= &  \lambda + \delta^{\topo}_{lb}(t+1)
	\end{flalign}
	The estimate error at the end of the planning horizon is:
	\begin{equation}
		\delta^{\topo}_{lb}(L-1) = \lambda
	\end{equation}
	Now, we can find out the estimation bound at the top of the given belief tree by calculating it from the bottom to the top, from $L-1$ to $0$.
\end{proof}

\section{Experiment Settings}
\label{sec:experiment-settings}
\subsection{Experiment 1}
\label{subsec:experiment-1-appendix}

In our current implementation, the belief tree $\BelTree^{\topo}$ that corresponds to an initial simplified topology $\topo$ is constructed by expanding the original observation space only at randomly chosen 15\% of the propagated belief nodes and switching the rest to an alternative observation space $\mathcal{O}=\mathcal{X}$ (see \eqref{equ:dirac-obs-model}), i.e.~considering the state space $\mathcal{X}$ instead of the observation space $\mathcal{Z}$. 
Then, at each iteration, if the condition \eqref{eq:sameaction} is not satisfied, we switch to a different (less simplified) topology $\topo'$ by turning $5$ randomly-chosen propagated belief nodes in $\topo$, that had an alternative observation space, back to the original observation space. 
Once \eqref{eq:sameaction} is satisfied, we are guaranteed to find the optimal action $a^{\star}_k$, and we terminate the process.

\subsection{Experiment 2}
\label{subsec:experiment-2-appendix}
The sparse sampling planner in a particle-belief POMDP setting has three parameters: planning horizon $d$, number of observation samples $K$, and number of the weighted state samples $N$ that represent the particle belief. In our experiments, we evaluate our approach and the original sparse sampling solver using two sets of these parameters: $d=3, K=50, N=50$ and $d=3, K=90, N=90$.

\subsection{Experiment 3}
\label{subsec:experiment-3-appendix}

The beacon navigation problem serves as a common benchmark for evaluating POMDP solvers \cite{LevYehudi24aaai, Barenboim23nips}.

The planning environment comprises a robot located at the initial belief, a beacon for generating observations, three obstacles, and a goal to reach. The locations of the robot, beacon, obstacles, and goal are defined in a 2D space as $\mathbf{x} \in \mathbb{R}^2$, $\mathbf{x}_b\in \mathbb{R}^2$, $\mathbf{x}_{o}\in \mathbb{R}^2$, and $\mathbf{x}_{g}\in \mathbb{R}^2$, respectively. The observation $\mathbf{z}\in \mathbb{R}^2$ is defined as the relative position with respect to the beacon, employing a Gaussian observation model defined as:
\begin{equation}
	P(\mathbf{z}|\mathbf{x}) = \begin{cases}
		\mathcal{N}(\mathbf{x}-\mathbf{x}_b,I_{2\times2}/(100||s-x_b||)), & \text{if } ||\mathbf{x}-\mathbf{x}_b||>1, \\
		\mathcal{N}(\mathbf{x}-\mathbf{x}_b,I_{2\times2}/100), & \text{otherwise}.
	\end{cases}
\end{equation}
The reward is assumed to be state-dependent: $r(b,a)=\emathbb{x|b}[r(x,a)]$.
The state-dependent reward function contains two parts, reward from reaching the goal $r_g$ and penalty from the obstacles $r_o$, as $r(\mathbf{x},\mathbf{a}) = r_g(\mathbf{x},\mathbf{a}) + r_o(\mathbf{x},\mathbf{a})$.
 The action $\mathbf{a}$ belongs to the action space $|\mathcal{A}|=\{[1.0, 0.0], [0.0, 1.0], [-1.0, 0.0], [0.0, -1.0]\}$ representing the movement of right, up, left, and down.
The reward of reaching the goal is defined as:
\begin{equation}
	r_g(\mathbf{x},\mathbf{a}) = \frac{50}{||\mathbf{x}-\mathbf{x}_{g}||+0.001} .
\end{equation}
The penalty from entering the nearby area of the three obstacles $o_1,o_2,o_3$ is defined as:
\begin{equation}
	r_o(\mathbf{x},\mathbf{a}) = -50 \text{, if }    ||\mathbf{x}-\mathbf{x}_{o_i}||\leq1, \forall i=1,2,3.
\end{equation}
The motion model follows a Gaussian distribution: 
\begin{equation}
	P(\mathbf{x}'|\mathbf{x},\mathbf{a}) = \mathcal{N}(\mathbf{x}+\mathbf{a}, \Sigma_T),
\end{equation}
with the covariance $\Sigma_T = I_{2\times2}/100$.

\begin{figure}[h!]
	\centering
	\includegraphics[width=0.6\textwidth]{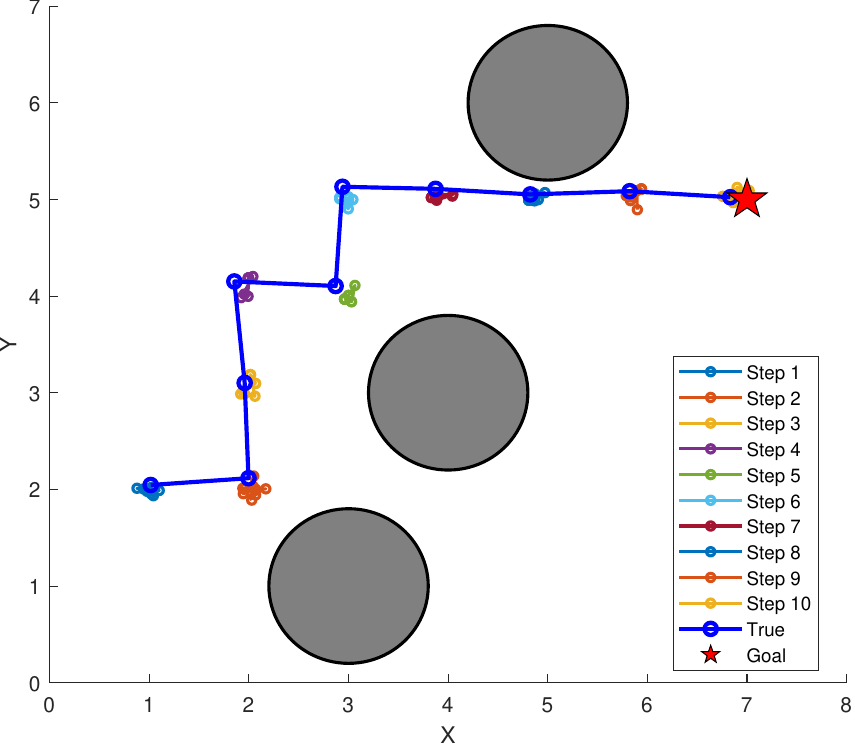}
	\vspace{-5pt}
	\caption{Simulation Trajectory of the Goal-Reaching Task }
	\label{fig:sim_traj}
\end{figure}

The trajectory of the goal-reaching tasks simulation is presented in Figure \ref{fig:sim_traj} below. The belief particles at each step are depicted by small dots of different colors. True positions of the robot are represented by blue dots and lines, while the gray circles indicate the obstacles with a penalty. The star symbolizes the goal to reach. This visualization serves to demonstrate the successful process by which our method reaches the goal. Our goal chooses the same optimal policy as the sparse sampling in the original full POMDP.
Table \ref{tab:traj_time} below demonstrates the total planning time for 10 steps until reach the goal using our proposed method and using sparse sampling in the original full problem. 
Own method achieves a significant speedup during the whole planning process. 

\begin{table}
	\centering
	\caption{Comparison of methods for an exact calculation of the Q function. 
	}
	\label{tab:traj_time}
	\resizebox{0.5\linewidth}{!}{
		\begin{tabular}{lcc}
			\hline
			Method &  Total Planning Time for 10 Steps (s) \\
			\hline
			Proposed  & \textbf{7.731} \\
			Full Problem  & 17.720 \\
			\hline
		\end{tabular}
	}
\end{table}
\subsection{Limitation}
 Our approach demonstrates effectiveness in scenarios characterized by a large observation space, which is consistent with the theoretical analysis of simplifying policy space. Despite requring multiple iterations, our simplification strategy is still able to reduce computation time.

 During the experiment, we also encountered failure cases in which nearly all nodes need to be un-simplified and use the original observation space so that the optimal action at the root could be distinguished.
 This implies that our method will eventually converge to a topology resembling the original one after numerous iterations, resulting in increased time costs.
 We address this dilemma as the trade-off between simplification and direct calculation, which all the simplification problems will face. 

 In addressing this dilemma, it is crucial to have a method that can intelligently and adaptively determine whether a situation is amenable to simplification.
 By evaluating the potential level of simplification before incurring high costs of explicit calculating, we can prune inadequate simplification candidates and further simplify the iteration process.
 We will focus on this problem in future work.

\end{document}


\title{No Correlations Involved: Decision Making Under Uncertainty in the Conservative Information Space \\  \textbf{Supplementary Material} \\ \vspace{15pt} \small{Technical Report ANPL-2016-01}}

\author{\renewcommand\footnotemark{}Vadim Indelman{*}%
\thanks{{*}Department of Aerospace Engineering, Technion - Israel Institute
of Technology, Haifa 32000, Israel.%
}}

\date{}

\maketitle
This document provides supplementary material to the paper \cite{Indelman16ral}. Therefore,
it should not be considered a self-contained document, but instead regarded as an appendix of \cite{Indelman16ral}, and cited as:

\vspace{10pt}

"V. Indelman, \emph{No Correlations Involved: Decision Making Under Uncertainty in 
a Conservative Sparse Information Space}, (Supplementary Material, ANPL-2016-01), IEEE Robotics and Automation Letters (RA-L),  accepted."

\vspace{10pt}

Throughout this report, standard notations are used to refer to equations from \cite{Indelman16ral} (e.g.~Eq.~(\ref{eq:ConservAppr})), while  equations introduced herein are represented by the corresponding Appendix letter, e.g.~Eqs.~(\ref{eq:R_plus}) and (\ref{eq:Det_Ic}). 

\bluecolor{
This document is organized as follows: Appendices A and B provide proofs for Lemmas \ref{lemma:detI} and \ref{lemma:detIc}; Appendix C proves Conjecture \ref{conj1-new} for two specific basic cases ($n=2$ and $n=3$); Appendix D provides additional numerical results, considering a high-dimensional decision making problem ($X\in \mathbb{R}^{1600}$).
}

\section*{Appendix A: Proof of Lemma \ref{lemma:detI}\label{sec:ProofLemmaDetI}}
\renewcommand\theequation{A\arabic{equation}}

Before presenting the proof of Lemma \ref{lemma:detI}, we recall
Givens rotations and introduce notations that will be used in the
proof.

Givens rotations is one possible approach to update an existing square
root information matrix $R\in\mathbb{R}^{n\times n}$ with the Jacobian
$A$, i.e.~to calculate the \emph{a posteriori} information matrix $R^{+}$:
\begin{equation}
\left[\begin{array}{c}
R\\
A
\end{array}\right]\rightarrow R^{+}\in\mathbb{R}^{n\times n}.
\label{eq:R_plus}
\end{equation}
One proceeds by applying Givens rotations to nullify all entries of
$A$ while the entries of $R$ are appropriately updated. To see that,
we consider the first Givens rotation: {\small{}
\begin{equation}
\left[\begin{array}{ccccc}
c &  &  &  & -s\\
\\
\\
\\
s &  &  &  & c
\end{array}\right]\left[\!\begin{array}{cccc}
r_{11} & r_{12} & \cdots & r_{1n}\\
 & r_{22} & \cdots & r_{2n}\\
 &  & \ddots & \vdots\\
 &  &  & r_{nn}\\
a_{1}^{\left(0\right)} & a_{2}^{\left(0\right)} & \vdots & a_{n}^{\left(0\right)}
\end{array}\!\right]\!=\!\left[\!\begin{array}{cccc}
{\color{blue}r_{11}^{+}} & {\color{blue}r_{12}^{+}} & \cdots & {\color{blue}r_{1n}^{+}}\\
 & r_{22} & \cdots & r_{2n}\\
 &  & \ddots & \vdots\\
 &  &  & r_{nn}\\
{\color{blue}0} & {\color{blue}a_{2}^{\left(1\right)}} & \vdots & {\color{blue}a_{n}^{\left(1\right)}}
\end{array}\!\right],
\end{equation}
}%
where the modified entries are denoted in blue, and $A\doteq\left[\begin{array}{ccc}
a_{1}^{\left(0\right)} & \cdots & a_{n}^{\left(0\right)}\end{array}\right]$. Since we consider unary observation models, and without loss of
generality, we arbitrarily assume the first state is measured, i.e.:
\begin{equation}
a_{1}^{\left(0\right)}\equiv a\ \ \text{and \ \ }a_{j}^{\left(0\right)}=0\ \ ,\ \ j>1.\label{eq:a_j0}
\end{equation}
We use the superscript to represent how many Givens rotations have
been performed thus far. It is not difficult to show that $s\doteq-\frac{a_{1}^{\left(0\right)}}{r_{11}^{+}}$
and $c\doteq\frac{r_{11}}{r_{11}^{+}}$, and hence: 
\begin{equation}
\left(r_{11}^{+}\right)^{2}=r_{11}^{2}+\left(a_{1}^{\left(0\right)}\right)^{2}\ \ ,\ \ a_{j}^{\left(1\right)}=\frac{-a_{1}^{\left(0\right)}r_{1j}+r_{11}a_{j}^{\left(0\right)}}{r_{11}^{+}}\label{eq:GivensFirst}
\end{equation}
for $1<j\leq n$. Note that although $a_{j}^{\left(0\right)}=0$ for
all $j$, $a_{j}^{\left(1\right)}\neq0$. However, all such entries
are \emph{proportional} to $a$: 
\begin{equation}
a_{j}^{\left(1\right)}=-a\frac{r_{1j}}{r_{11}^{+}}.\label{eq:a_j1}
\end{equation}
This fact will be used in the sequel.

Consider now the $i$th application of Givens rotation, that nullifies
$a_{i}^{\left(i-1\right)}$ and modifies additional entries as shown
below. {
\begin{multline}
\left[\begin{array}{ccccccc}
r_{11}^{+} & \!\cdots\! & r_{1,i-1}^{+} & r_{1,i}^{+} & r_{1,i+1}^{+} & \!\cdots\! & r_{1n}^{+}\\
 & \!\ddots\! & \vdots & \vdots & \vdots & \ddots & \vdots\\
 &  & r_{i-1,i-1}^{+} & r_{i-1,i}^{+} & r_{i-1,i+1}^{+} & \!\cdots\! & r_{i-1,n}^{+}\\
 &  &  & r_{ii} & r_{i,i+1} & \!\cdots\! & r_{in}\\
 &  &  &  & r_{i+1,i+1} & \!\cdots\! & r_{i+1,n}\\
 &  &  &  &  & \!\ddots\! & \vdots\\
 &  &  &  &  &  & r_{nn}\\
0 & \!\cdots\! & 0 & a_{i}^{\left(i-1\right)} & a_{i+1}^{\left(i-1\right)} & \!\cdots\! & a_{n}^{\left(i-1\right)}
\end{array}\right]\! \rightarrow \\ 
\rightarrow \!\left[\begin{array}{ccccccc}
r_{11}^{+} & \!\cdots\! & r_{1,i-1}^{+} & r_{1,i}^{+} & r_{1,i+1}^{+} & \!\cdots\! & r_{1n}^{+}\\
 & \!\ddots\! & \vdots & \vdots & \vdots & \!\ddots\! & \vdots\\
 &  & r_{i-1,i-1}^{+} & r_{i-1,i}^{+} & r_{i-1,i+1}^{+} & \!\cdots\! & r_{i-1,n}^{+}\\
 &  &  & {\color{blue}r_{ii}^{+}} & {\color{blue}r_{i,i+1}^{+}} & \!\cdots\! & {\color{blue}r_{in}^{+}}\\
 &  &  &  & r_{i+1,i+1} & \!\cdots\! & r_{i+1,n}\\
 &  &  &  &  & \!\ddots\! & \vdots\\
 &  &  &  &  &  & r_{nn}\\
0 & \!\cdots\! & 0 & {\color{blue}0} & {\color{blue}a_{i+1}^{\left(i\right)}} & {\color{blue}\!\cdots\!} & {\color{blue}a_{n}^{\left(i\right)}}
\end{array}\right]
\end{multline}
}{\small \par}

{\small{}}%

The following expressions for $r_{ii}^+$ and $a_j^{(i)}$ can be obtained by generalizing Eq.~(\ref{eq:GivensFirst}):
\begin{equation}
\left(r_{ii}^{+}\right)^{2}=r_{ii}^{2}+\left(a_{i}^{\left(i-1\right)}\right)^{2}\ \ ,\ \ a_{j}^{\left(i\right)}=\frac{-a_{i}^{\left(i-1\right)}r_{ij}+r_{ii}a_{j}^{\left(i-1\right)}}{r_{ii}^{+}},\label{eq:Givens_i}
\end{equation}
with $i<j\leq n$, and, of course, $a_{i}^{\left(i\right)}=0$.
Recall that 
\begin{equation}
\left|\Lambda^{+}\right|=\prod_{i=1}^{n}\left(r_{i,i}^{+}\right)^{2}.
\end{equation}
We now prove Lemma \ref{lemma:detI} using mathematical induction.

\paragraph*{Basis}

We show Lemma \ref{lemma:detI} holds for $n=2$ (and $n=1$). Using
Eq.~(\ref{eq:Givens_i}), $\left|\Lambda^{+}\right|$ can be written as
\begin{multline}
\left|\Lambda^{+}\right|=\left(r_{1,1}^{+}\right)^{2}\left(r_{2,2}^{+}\right)^{2}=\left(r_{11}^{+}\right)^{2}\left[r_{22}^{2}+\left(a_{2}^{\left(1\right)}\right)^{2}\right]=\\
\left(r_{11}^{+}\right)^{2}\frac{r_{22}^{2}\left(r_{11}^{+}\right)^{2}+\left(-a_{1}^{\left(0\right)}r_{12}+r_{11}a_{2}^{\left(0\right)}\right)^{2}}{\left(r_{11}^{+}\right)^{2}}
\end{multline}
Recalling Eq.~(\ref{eq:a_j0}) we get
\begin{equation}
\left|\Lambda^{+}\right|=r_{11}^{2}r_{22}^{2}+a^{2}\left(r_{22}^{2}+r_{12}^{2}\right)=\eta_{2}+a^{2}\gamma_{2}\left(R\right)
\end{equation}
with $\gamma_{2}\left(R\right)>0$.

\paragraph*{Inductive step}

Consider Lemma \ref{lemma:detI} holds for $n=k$:
\begin{equation}
\left|\Lambda^{+}\right|=\prod_{i=1}^{n=k}\left(r_{i,i}^{+}\right)^{2}=\eta_{k}+a^{2}\gamma_{k}\left(R\right),\label{eq:lemma_k}
\end{equation}
and $\gamma_{k}\left(R\right)>0$. Letting$\left|\Lambda_{k}^{+}\right|\doteq\prod_{i=1}^{k}\left(r_{i,i}^{+}\right)^{2},$
Eq.~(\ref{eq:lemma_k}) corresponds to

\begin{equation}
\left(r_{kk}^{+}\right)^{2}=\frac{1}{\left|\Lambda_{k-1}^{+}\right|}\left[\eta_{k}+a^{2}\gamma_{k}\left(R\right)\right].\label{eq:r_kk}
\end{equation}
We now prove Lemma \ref{lemma:detI} is satisfied also for $n=k+1$,
i.e.: 
\begin{equation}
\left|\Lambda_{k+1}^{+}\right|=\eta_{k+1}+a^{2}\gamma_{k+1}\left(R\right).
\end{equation}
We start with deriving an expression for $\left(r_{k+1,k+1}^{+}\right)^{2}=r_{k+1,k+1}^{2}+\left(a_{k+1}^{\left(k\right)}\right)^{2}$:
\begin{multline}
\left(\! r_{k+1,k+1}^{+}\!\right)^{2}\!=\! r_{k+1,k+1}^{2}\!+\!\frac{\left(-a_{k}^{\left(k-1\right)}r_{k,k+1}\!+\! r_{k,k}a_{k+1}^{\left(k-1\right)}\right)^{2}}{\left(r_{k,k}^{+}\right)^{2}}\!=\!\\
\frac{1}{\left(r_{k,k}^{+}\right)^{2}}\!\left[\! r_{k+1,k+1}^{2}\left(r_{k,k}^{+}\right)^{2}\!+\!\left(-a_{k}^{\left(k-1\right)}r_{k,k+1}\!+\! r_{k,k}a_{k+1}^{\left(k-1\right)}\right)^{2}\!\right] 
\end{multline}
Plugging in the expression for $\left(r_{kk}^{+}\right)^{2}$ from
Eq.~(\ref{eq:r_kk}) yields
\begin{multline} 
\left(r_{k+1,k+1}^{+}\right)^{2}=\frac{1}{\left(r_{k,k}^{+}\right)^{2}}\left[\frac{r_{k+1,k+1}^{2}}{\left|\Lambda_{k-1}^{+}\right|}\left[\eta_{k}+a^{2}\gamma_{k}\left(R\right)\right]+
\right.  \\
\left. \left(-a_{k}^{\left(k-1\right)}r_{k,k+1}+r_{k,k}a_{k+1}^{\left(k-1\right)}\right)^{2}\right] \label{eq:r_k_plus_1_a}
\end{multline} 
Now, it is not difficult to show that by recursively using the relations
(\ref{eq:Givens_i}) we can express $a_{k}^{\left(k-1\right)}$ in
terms of $a_{k}^{\left(1\right)}$. Recalling Eq.~(\ref{eq:a_j1})
we get 
\begin{equation}
a_{k}^{\left(k-1\right)}\!=\!\frac{\!-\! a_{k-1}^{\left(k-2\right)}r_{k-1,k}\!+\! r_{k-1,k-1}a_{k}^{\left(k-2\right)}}{r_{k-1,k-1}^{+}}\!=\!\cdots\!=\! a\frac{f_{k}^{\left(k-1\right)}\!\left(R\right)}{\sqrt{\left|\Lambda_{k-1}^{+}\right|}},\label{eq:a_k}
\end{equation}
and similarly for $a_{k+1}^{\left(k-1\right)}$
\begin{equation}
a_{k+1}^{\left(k-1\right)}=\cdots=a\frac{f_{k+1}^{\left(k-1\right)}\left(R\right)}{\sqrt{\left|\Lambda_{k-1}^{+}\right|}},\label{eq:a_k_1}
\end{equation}
where $f_{k}^{\left(k-1\right)}\left(R\right)$ and $f_{k+1}^{\left(k-1\right)}\left(R\right)$
are only functions of entries of $R$. We stress that this statement
is valid only because of Eq.~(\ref{eq:a_j1}), which corresponds
to assuming unary measurement models. 

Substituting expressions (\ref{eq:a_k}) and (\ref{eq:a_k_1}) into
Eq.~(\ref{eq:r_k_plus_1_a}) yields 
\begin{equation}
\left(r_{k+1,k+1}^{+}\right)^{2}=\frac{1}{\left|\Lambda_{k}^{+}\right|}\left[\eta_{k+1}+a^{2}\gamma_{k+1}\left(R\right)\right],\label{eq:r_k_pls_1}
\end{equation}
with 
\begin{equation}
\gamma_{k+1}\left(R\right)\doteq r_{k+1,k+1}^{2}\gamma_{k}\left(R\right)+\left(-f_{k}^{\left(k-1\right)}\left(R\right)r_{k,k+1}+f_{k+1}^{\left(k-1\right)}\left(R\right)r_{k,k}\right)^{2}.
\end{equation}
Since $\left|\Lambda_{k+1}^{+}\right|=\left|\Lambda_{k}^{+}\right|\cdot\left(r_{k+1,k+1}^{+}\right)^{2}$,
we get
\begin{equation}
\left|\Lambda_{k+1}^{+}\right|=\eta_{k+1}+a^{2}\gamma_{k+1}\left(R\right).
\end{equation}
We showed Lemma \ref{lemma:detI} holds for both the basis and inductive
steps; hence, according to mathematical induction it holds for all
natural $n$.
\hfill $\blacksquare$

\section*{Appendix B: Proof of Lemma \ref{lemma:detIc}\label{sec:ProofLemmaDetIc}}
\renewcommand\theequation{B\arabic{equation}}

Recall Eq.~(\ref{eq:DetIc}): $\left|\Lambda_{c}^{+}\right|=\prod_{i=1}^{n}\left(r_{c,ii}^{+}\right)^{2}$.
Since $R_{c}$ is diagonal, it is not difficult to show that (see
Eq.~(\ref{eq:Givens_i})) $r_{c,ii}^{+}=r_{c,ii}$ for $i>1$, i.e.
only the upper left entry in matrix $R_{c}$ is actually updated due
to Jacobian $A$. Thus, we can write:

\begin{equation}
\left|\Lambda_{c}^{+}\right|=\left(r_{c,11}^{2}+a^{2}\right)\prod_{i=2}^{n}r_{c,ii}^{2}.\label{eq:Det_Ic}
\end{equation}
According to Eq.~(\ref{eq:ConservAppr}), $r_{c,ii}^{2}=w_{i}\Sigma_{ii}^{-1}$
where $\Sigma_{ii}$ is the corresponding entry on the diagonal of
the covariance matrix $\Sigma\equiv \Lambda^{-1}$. Assuming, for simplicity
$w_{i}=w=1/n$, Eq.~(\ref{eq:Det_Ic}) turns into 
\begin{equation}
\left|\Lambda_{c}^{+}\right|=\left(\Sigma_{11}^{-1}+na^{2}\right)n^{-n}\prod_{i=2}^{n}\Sigma_{ii}^{-1}.\label{eq:Det_Ic-1}
\end{equation}
 In practice, as detailed in \cite{Golub80laa} (see also \cite{Kaess09ras}),
calculation of $\Sigma_{ii}$ can be efficiently performed directly
from the nonzero entries of $R$, without the need in calculating
an inverse of a large matrix: 
\begin{equation}
\Sigma_{ll}=\frac{1}{r_{ll}}\left(\frac{1}{r_{ll}}-\sum_{j=l+1}^{n}r_{lj}\Sigma_{jl}\right)\label{eq:Sigma_ll}
\end{equation}
\begin{equation}
\Sigma_{il}=\frac{1}{r_{ii}}\left(-\sum_{j=i+1}^{l}r_{ij}\Sigma_{jl}-\sum_{j=l+1}^{n}r_{ij}\Sigma_{lj}\right)\label{eq:Sigma_il}
\end{equation}
Based on Eqs.~(\ref{eq:Sigma_ll})-(\ref{eq:Sigma_il}), it is possible
to show that $\Sigma_{ii}$ can be written, for all $i$, as 
\begin{equation}
\Sigma_{ii}=\frac{\tilde{\gamma}_{i}}{\prod_{j=i}^{n}r_{jj}^{2}},\label{eq:Sigma_ii_1}
\end{equation}
where $\tilde{\gamma}$ is only a function of elements of $R$. Recalling
the definition of $\eta_{i}$ from Eq.~(\ref{eq:eta}), we can write 
\begin{equation}
\prod_{j=i}^{n}r_{jj}^{2}=\prod_{j=1}^{n}r_{jj}^{2}\diagup\prod_{j=1}^{i-1}r_{jj}^{2}=\frac{\eta_{n}}{\eta_{i-1}},
\end{equation}
with the convention that $\eta_{0}\doteq1$. Denoting, for convenience
$\gamma_{c,n-i+1}\doteq\tilde{\gamma}_{i}$, Eq.~(\ref{eq:Sigma_ii_1})
can be rewritten as
\begin{equation}
\Sigma_{ii}=\gamma_{c,n-i+1}\frac{\eta_{i-1}}{\eta_{n}}.\label{eq:Sigma_ii}
\end{equation}
Substituting Eq.~(\ref{eq:Sigma_ii}) into Eq.~(\ref{eq:Det_Ic-1})
results in
\begin{equation}
\left|\Lambda_{c}^{+}\right|=\left(\eta_{n}+na^{2}\gamma_{c,n}\right)n^{-n}\frac{\prod_{i=2}^{n} \frac{\eta_{n}}{\eta_{i-1}}}{\prod_{i=1}^{n}\gamma_{c,n-i+1}}.\label{eq:Icplus}
\end{equation}
Taking a closer look at $\prod_{i=2}^{n}\frac{\eta_{n}}{\eta_{i-1}}$
we can see that
\begin{alignat}{1}
\prod_{i=2}^{n}\frac{\eta_{n}}{\eta_{i-1}} & =\left(r_{22}^{2}r_{33}^{2}\cdots r_{nn}^{2}\right)\left(r_{33}^{2}\cdots r_{nn}^{2}\right)\cdots\left(r_{n-1,n-1}^{2}r_{nn}^{2}\right)r_{nn}^{2}\nonumber \\
 & =\prod_{i=2}^{n}r_{ii}^{2\left(i-1\right)}\equiv\alpha_{n}/n^{-n},
\end{alignat}
where $\alpha_{n}$ is defined in Eq.~(\ref{eq:alpha}). Defining
$\beta_{n}$ as
\begin{equation}
\beta_{n}\doteq\prod_{i=1}^{n}\gamma_{c,n-i+1},
\end{equation}
Eq.~(\ref{eq:Icplus}) can be finally rewritten as 
\begin{equation}
\left|\Lambda_{c}^{+}\right|=\frac{\alpha_{n}}{\beta_{n}}\left[\eta_{n}+na^{2}\gamma_{c,n}\right].
\end{equation}
This completes the proof of Lemma \ref{lemma:detIc}. \hfill $\blacksquare$


\section*{Appendix C}
\renewcommand\theequation{C\arabic{equation}}

In this appendix, we prove the relation from \bluecolor{Conjecture \ref{conj1-new}}
\begin{equation}
\left|\Lambda_{c}^{+}\right|=\frac{\alpha_{n}}{\beta_{n}}\left[n\left|\Lambda^{+}\right|-\left(n-1\right)\eta_{n}\right],
\label{eq:Proposition-appendix}
\end{equation}
for $n=2$ and $n=3$, thereby proving Conjectures \ref{conj1-gamma} and \ref{conj1-new} for these cases. In the following we use, for simplicity, $w_{i}=w=1/n$.

We start with $n=2$. According to Lemma \ref{lemma:detI} it is possible
to show that
\begin{equation}
\left|\Lambda^{+}\right|=\prod_{i=1}^{2}\left(r_{ii}^{+}\right)^{2}=a_{1}^{2}\left(r_{12}^{2}+r_{22}^{2}\right)+r_{11}^{2}r_{22}^{2}.
\end{equation}
On the other hand, using Eqs.~(\ref{eq:Sigma_ll})-(\ref{eq:Sigma_il})
we get 
\begin{equation}
r_{c11}^{2}=\frac{1}{2}\frac{r_{11}^{2}r_{22}^{2}}{r_{12}^{2}+r_{22}^{2}}\ \ ,\ \ r_{c22}^{2}=\frac{1}{2}r_{22}^{2}.\label{eq:ConservativeN2-2}
\end{equation}
Recalling Eq.~(\ref{eq:Det_Ic}), $\left|\Lambda_{c}^{+}\right|=\left(r_{c11}^{2}+a_{1}^{2}\right)r_{c22}^{2}$,
and substituting Eq.~(\ref{eq:ConservativeN2-2}) we obtain the following
recursive relation in $\left|\Lambda^{+}\right|$:
\begin{equation}
\left|\Lambda_{c}^{+}\right|=\frac{1}{2^{2}}r_{22}^{2}\frac{2\left|\Lambda^{+}\right|-r_{11}^{2}r_{22}^{2}}{r_{12}^{2}+r_{22}^{2}}.
\end{equation}
This expression indeed corresponds to Eq.~(\ref{eq:Proposition-appendix}).

Considering now $n=3$, it is possible to show that
\begin{equation}
\left|\Lambda^{+}\right|\!=\! r_{11}^{2}r_{22}^{2}r_{33}^{2}+a_{1}^{2}\left[r_{33}^{2}\left(r_{12}^{2}+r_{22}^{2}\right)\!+\!\left(r_{12}r_{23}-r_{13}r_{22}\right)^{2}\right].\label{eq:Iplus-1-1}
\end{equation}
Similarly, to the previous case, using Eqs.~(\ref{eq:Sigma_ll})-(\ref{eq:Sigma_il})
we get $r_{c22}^{2}=\frac{1}{3}\frac{r_{22}^{2}r_{33}^{2}}{r_{23}^{2}+r_{33}^{2}}$,
$r_{c33}^{2}=\frac{1}{3}r_{33}^{2}$ and
\begin{equation}
r_{c11}^{2}=\frac{1}{3}\frac{r_{11}^{2}r_{22}^{2}r_{33}^{2}}{r_{33}^{2}\left(r_{12}^{2}+r_{22}^{2}\right)+\left(r_{12}r_{23}-r_{13}r_{22}\right)^{2}}.\label{eq:ConservativeN3_b-2}
\end{equation}
Noting that $\left|\Lambda_{c}^{+}\right|=\left(r_{c11}^{2}+a_{1}^{2}\right)r_{c22}^{2}r_{c33}^{2}$
and substituting the above relations it is not difficult to show that
\begin{equation}
\left|\Lambda_{c}^{+}\right|=\frac{1}{3^{3}}\frac{r_{22}^{2}r_{33}^{4}}{r_{23}^{2}+r_{33}^{2}}\left(\frac{3\left|\Lambda^{+}\right|-2r_{11}^{2}r_{22}^{2}r_{33}^{2}}{r_{33}^{2}\left(r_{12}^{2}+r_{22}^{2}\right)+\left(r_{12}r_{23}-r_{13}r_{22}\right)^{2}}\right).
\end{equation}
As previously, this expression indeed corresponds to Eq.~(\ref{eq:Proposition-appendix}).


\bluecolor{
\section*{Appendix D: Additional Results}
\renewcommand\theequation{D\arabic{equation}}

In this appendix we demonstrate the proposed concept in a larger sensor-deployment problem than the one considered in \cite{Indelman16ral}. Specifically, we have a $40\times40$ grid (instead of $10\times10$ as in \cite{Indelman16ral}), which corresponds to $X\in \mathbb{R}^{1600}$ and $\Sigma\in \mathbb{R}^{1600\times1600}$. Figure \ref{fig:UncertaintyFilerLarge} provides the results: the prior uncertainty field is shown in Figure \ref{fig:UncertaintyFilerLarge-field}, the corresponding running time for making $10$ greedy sensor deployment decisions using the original and conservative information space is shown in Figure \ref{fig:UncertaintyFilerLarge-timing}. Figures (\ref{fig:UncertaintyFilerLarge-dec})-(\ref{fig:UncertaintyFilerLarge-dec-sorted-sep}) show the impact of candidate actions is preserved, as stated by Conjecture \ref{cnj-high-dim}. While this is not easily inferred from Figure \ref{fig:UncertaintyFilerLarge-dec}, we provide a zoom-in in Figure \ref{fig:UncertaintyFilerLarge-dec-zoom}. We also show in Figures \ref{fig:UncertaintyFilerLarge-dec-sorted} and \ref{fig:UncertaintyFilerLarge-dec-sorted-sep} the impact of candidate actions when sorting the x-axis (i.e.~candidate actions) considering the objective function\footnote{We use the logarithm, a monotonic function, for numerical reasons.} $log(det(\Lambda^+))$ that uses the original information space. The same ordering is then used for the conservative information space. Thus, if the trend was different, the  resulting curve for the conservative case would not be monotonically decreasing.

\begin{figure}
\subfloat[\label{fig:UncertaintyFilerLarge-field}]{
	\includegraphics[width=0.5\columnwidth]{Figures/SensorDeployment/LargeExample/00000_UncertainField.eps}}
\subfloat[\label{fig:UncertaintyFilerLarge-timing}]{
\includegraphics[width=0.5\columnwidth]{Figures/SensorDeployment/LargeExample/Timing.eps}}

\subfloat[\label{fig:UncertaintyFilerLarge-dec}]{
	\includegraphics[width=0.5\columnwidth]{Figures/SensorDeployment/LargeExample/Decisions.eps}}
\subfloat[\label{fig:UncertaintyFilerLarge-dec-zoom}]{
\includegraphics[width=0.5\columnwidth]{Figures/SensorDeployment/LargeExample/Decisions_zoom.eps}}

\subfloat[\label{fig:UncertaintyFilerLarge-dec-sorted}]{
	\includegraphics[width=0.5\columnwidth]{Figures/SensorDeployment/LargeExample/Decisions_sorted.eps}}
\subfloat[\label{fig:UncertaintyFilerLarge-dec-sorted-sep}]{
\includegraphics[width=0.5\columnwidth]{Figures/SensorDeployment/LargeExample/Decisions_sorted_sep.eps}}

\caption{\label{fig:UncertaintyFilerLarge}\bluecolor{Uncertainty field synthetic example: (a) A priori variance in each cell of the $N \times N$ grid with $N=40$, which corresponds to $X\in \mathbb{R}^{1600}$ and $\Sigma\in \mathbb{R}^{1600\times1600}$; (b) Timing results for making $10$ sequential greedy decisions using the original and conservative information space. (c) Impact of each candidate decision (sensor location) using the original and conservative information matrices. A zoom-in is shown in (d). Although values are different, the trend is identical in both cases for any two candidate actions, as stated by Conjecture \ref{cnj-high-dim}. (e) Impact of candidate decisions from (c), with both curves sorted according to $log(det(\Lambda^+))$. The monotonically decreasing curve for $log(det(\Lambda^+_c))$ indicates an identical trend in both cases. (d) Numerical values of each curve from (e).}}
\end{figure}

}

\bibliographystyle{IEEEtran}
\bibliography{../../../../../../refs}